\pdfoutput=1
\documentclass[11pt,a4paper]{article}
\usepackage[utf8]{inputenc}
\usepackage[T1]{fontenc}
\usepackage{mathptmx}
\usepackage[scaled=0.92]{helvet}
\usepackage{courier}
\usepackage[margin=2.5cm]{geometry}
\usepackage{amsmath,amssymb}
\usepackage{graphicx}
\usepackage{booktabs}
\usepackage{array}
\usepackage{tabularx}
\usepackage{placeins}
\usepackage[round]{natbib}
\usepackage[font=small,labelfont=bf,width=0.92\textwidth]{caption}
\usepackage{microtype}
\usepackage[colorlinks=true,linkcolor=blue!60!black,citecolor=blue!60!black,urlcolor=blue!60!black]{hyperref}
\usepackage{xcolor}
\newcolumntype{Y}{>{\raggedright\arraybackslash}X}
\newcolumntype{P}[1]{>{\raggedright\arraybackslash}p{#1}}
\title{From 12 to 200 Demonstrations: How ACT, Diffusion Policy, SmolVLA and \texorpdfstring{$\pi_{0.5}$}{pi0.5} Scale on a Low-Cost Arm}
\author{Simon Cowling\\ fysical.ai, Denmark\\ \texttt{simon@fysical.ai}}
\date{}

\begin{document}
\maketitle

\begin{abstract}
We trained four widely used imitation-learning policies, ACT, Diffusion Policy, SmolVLA and $\pi_{0.5}$, on nested subsets of 12, 25, 50, 100 and 200 teleoperated demonstrations of one pick-and-place task with distractor objects, using one fixed training recipe per policy. Every trained model was then run on a physical SO-101 arm on a fixed set of object placements: 1,340 scored episodes in 126 blocks.

ACT, trained from scratch, rose from 1 success in 40 episodes at 12 demonstrations to 37 in 40 at 100. $\pi_{0.5}$ with its action expert fine-tuned was the best policy when data was scarce (16 of 40 at 12 demonstrations, 28 of 40 at 25) and then stayed near 70\%; from 100 demonstrations ACT was ahead, by a margin this study does not fully resolve. SmolVLA and Diffusion Policy peaked near 50\%, and both were affected by a failure to start: SmolVLA's arm never started moving in 47\% of episodes, and Diffusion Policy hardly ever started until an idle lead-in was trimmed from the demonstrations. For $\pi_{0.5}$ the fine-tuning recipe mattered more than the amount of data: low-rank adaptation succeeded in 4 of 110 episodes, against 130 of 200 for action-expert fine-tuning.

From the joint and camera recordings we place every episode on a four-stage ladder (started, reached the ball, carried it, delivered it) and show that for ACT more data moves failures from reaching, to grasping, to none, while the failures $\pi_{0.5}$ retains are in grasping. We also measure how weak a ten-episode block is as a measurement (the same checkpoint tested twice differs by 2 successes in 10 on average), separate seed-to-seed variation from that noise, and audit the human scoring against the recordings. Training length was fixed within each recipe, so the amount of data and the number of times it was seen change together; we state which conclusions survive that confound and which do not.

\end{abstract}
\FloatBarrier
\section{Introduction}\label{sec:1}
Low-cost arms and open-source software have made real-robot imitation learning accessible to small laboratories and individuals. A SO-101 arm costs a few hundred dollars in parts \citep{knight2024so101}, and the LeRobot library \citep{cadene2024lerobot,cadene2026lerobot} provides recording, training and evaluation for a range of policies behind one interface. Someone starting a new task therefore faces two practical questions at once: which policy to train, and how many demonstrations to record.

The literature does not answer either question well for this setting. Each policy was introduced on different robots, tasks and data volumes. ACT was shown with 50 demonstrations per task on a bimanual rig \citep{zhao2023act}, Diffusion Policy on a different set of simulated and real tasks \citep{chi2023diffusion}, and the vision-language-action models SmolVLA \citep{shukor2025smolvla} and $\pi_{0.5}$ \citep{black2025pi05} after pretraining on large multi-robot datasets. Pretrained models are expected to need fewer task demonstrations; models trained from scratch are cheaper to train and run. Recent benchmarks on low-cost arms compare several of these policies at one data size, or at a few sizes with twenty trials per model (Section~\ref{sec:2}), which is too few to say which differences are real.

This paper reports a controlled study of that question on one physical task. We recorded 225 teleoperated demonstrations of a pick-and-place task with two distractor objects, fixed nested subsets of 12, 25, 50, 100 and 200 demonstrations, and trained ACT, Diffusion Policy, SmolVLA and $\pi_{0.5}$ on each subset with several training seeds. Every trained model was run on the arm on ten fixed object placements under a 30-second limit. In total we scored 1,340 episodes in 126 blocks over 17 days, and recorded joint states, commanded actions and both camera streams for every episode.

The contributions are:

\begin{enumerate}
\item \textbf{Scaling curves for four policies on one physical task.} Success against number of demonstrations from 12 to 200, with confidence intervals and a trend estimate for each policy, including a reversal of order between ACT trained from scratch and fine-tuned $\pi_{0.5}$ (Section~\ref{sec:4.1}).
\item \textbf{A stage-by-stage account of failure taken from the recordings.} Each episode is placed on a four-stage ladder using joint and wrist-camera data, which shows where each policy loses episodes and how that shifts with data (Section~\ref{sec:4.2}).
\item \textbf{Initiation failure.} Two of the four policies often never start moving. For Diffusion Policy we show by intervention on the training data that removing an idle lead-in from the start of the demonstrations removes the failure (Section~\ref{sec:4.3}).
\item \textbf{Recipe sensitivity of $\pi_{0.5}$.} Two fine-tuning recipes that differ in which weights are trained give 4\% and 65\% success across the same data budgets, a larger effect than the amount of data had for any policy except ACT (Section~\ref{sec:4.4}).
\item \textbf{How much a block of ten episodes can tell you.} Re-tests of the same checkpoint, a separation of seed-to-seed variation from measurement noise, and an audit of the operator's scoring against the recordings (Sections~\ref{sec:3.6} and~\ref{sec:4.5}).
\end{enumerate}
\subsection{Scope of claims}\label{sec:1.1}
The study has a narrow base and one structural confound, and both shape how the results should be read. We state them here; Section~\ref{sec:7} gives the full list of limitations.

\begin{itemize}
\item \textbf{One task, one arm, one operator.} The rankings are for this task and setup. We make no claim that they transfer to other tasks.
\item \textbf{Recipes, not architectures.} Each policy was trained with one fixed recipe, close to the library's documented defaults. The comparison is between policies as a practitioner would typically run them, not between architectures at their best.
\item \textbf{Fixed training length.} The number of training steps was the same at every data budget, so a model trained on 200 demonstrations saw each frame about 17 times less often than one trained on 12. Where a curve flattens or falls at 100 and 200 demonstrations, this study cannot tell a ceiling from under-training (Section~\ref{sec:5}).
\item \textbf{Testing was dimmer than training.} Main-comparison blocks ran at roughly half the illuminance of the demonstrations. All policies faced this shift, though not to an identical degree.
\end{itemize}
\FloatBarrier
\section{Related work}\label{sec:2}
\paragraph{The four policies.} ACT \citep{zhao2023act} is a transformer trained from scratch as a conditional variational autoencoder that predicts a chunk of future actions. Diffusion Policy \citep{chi2023diffusion} generates an action sequence by iterative denoising. SmolVLA \citep{shukor2025smolvla} is a compact vision-language-action model that pairs the SmolVLM backbone \citep{marafioti2025smolvlm} with a flow-matching action expert and is pretrained on community-collected LeRobot datasets. $\pi_0$ \citep{black2025pi0} and $\pi_{0.5}$ \citep{black2025pi05} pair a PaliGemma backbone \citep{beyer2024paligemma} with a flow-matching action expert and are pretrained on a much larger multi-robot corpus. All four predict action chunks, and all four are available in LeRobot, which is what makes a like-for-like comparison practical.

\paragraph{Fine-tuning large policies.} A pretrained vision-language-action model can be adapted by training all weights, by training low-rank adapters \citep[LoRA;][]{hu2022lora}, or by training the action expert while the vision-language backbone is frozen or shielded from its gradients \citep{driess2025insulating}. The openpi reference code publishes full fine-tuning configurations for $\pi_{0.5}$ and low-memory LoRA configurations for $\pi_0$ only \citep{openpi2025}. We are not aware of a published real-robot comparison of these options for $\pi_{0.5}$ across data budgets.

\paragraph{How success scales with demonstrations.} \citet{mandlekar2021robomimic} varied dataset size among other factors for several offline learning algorithms, mostly in simulation. \citet{lin2025scaling} found that generalisation follows an approximate power law in the number of training environments and objects, and that diversity matters more than the number of demonstrations in each. \citet{brohan2023rt1} reported the effect of data size and diversity for RT-1. \citet{black2025pi0} fine-tuned $\pi_0$ with different amounts of task data against ACT and Diffusion Policy trained from scratch, and \citet{zhang2025experiences} found ACT the most stable of four policies on in-distribution tasks and $\pi_0$ the most adaptable outside them. None of these traces a per-policy curve on one physical task with enough repeated trials to put an interval on each point.

\paragraph{Benchmarks on low-cost arms.} Three recent preprints are closest to this work. \citet{wang2026surgery} compare ACT, Diffusion Policy, SmolVLA and $\pi_0$ on a suture-following task with 160 demonstrations on an open-source arm, with dataset size as one of three evaluation dimensions, and find $\pi_0$ the strongest and most data-efficient. \citet{yu2026benchmarking} benchmark $\pi_{0.5}$, SmolVLA, Wall-X and ACT on four SO-101 tasks with a failure taxonomy, and find that the pretrained models generally outperform ACT. \citet{selvaraj2026armnetbench} run seven policies on twelve tasks with 50 demonstrations each across a fleet of SO-101 cells. Our study is narrower and deeper: one task, a 17-fold range of data, repeated training seeds and repeated tests of the same checkpoint, and an analysis of every episode's recording. It also reaches a different conclusion at the high-data end, where the policy trained from scratch does at least as well as the pretrained one.

\paragraph{Idle actions.} Demonstrations often contain stretches in which the robot does not move, and policies can learn to reproduce them. \citet{chi2023diffusion} discuss robustness to idle actions and argue that predicting action sequences reduces the problem. \citet{kim2024openvla} report that OpenVLA froze during evaluation until near-zero actions were filtered from the training data, and the openpi code filters idle transitions from DROID \citep{openpi2025}. \citet{chen2025idling} study policy idling directly and propose perturbations to escape it. Our contribution to this line is a clean intervention on one dataset: the same demonstrations with and without their idle lead-in, for a policy that predicts action sequences.

\paragraph{Evaluation practice.} \citet{kressgazit2024empirical} set out best practices for evaluating learned robot policies, including reporting conditions, failure modes and statistical uncertainty. \citet{vincent2024generalizable} and \citet{snyder2025step,snyder2026beyond} give methods for bounding and comparing success rates from few trials, and the \citet{tri2025lbm} applies such methods at scale. In reinforcement learning, \citet{henderson2018matters} and \citet{agarwal2021precipice} showed how seed variance and small numbers of runs distort conclusions. We add two measurements that are rarely reported for real-robot studies: how often the same checkpoint gives a different outcome on the same placement, and how often a human scorer's sheet disagrees with the recording.

\FloatBarrier
\section{Experimental setup}\label{sec:3}
One task was recorded once, subsets of the recording were fixed once, and every model was trained and tested by the same procedure. This section describes each step in the order it happened.

\subsection{Hardware and workspace}\label{sec:3.1}
Table~\ref{tab:hw} lists the hardware, software and workspace.

\begin{table}[t]
\centering
\small
\caption{\textbf{Hardware, software and workspace.}}
\label{tab:hw}
\begin{tabular}{P{6.66cm}P{8.78cm}}
\toprule
Component & Description \\
\midrule
Follower arm & SO-101 follower \citep{knight2024so101}: five arm joints and a gripper, six Feetech STS3215 bus servos. Driven through LeRobot's \texttt{so101\_follower} with joint angles in degrees and the gripper on a 0 to 100 scale. Silicone tape was added to the inner gripper surfaces to increase friction. The follower arm was mounted on a custom raised platform to simulate mounting on a lerobot Kiwi. \\
Leader arm & SO-101 leader used for teleoperation (mounted on the same type of platform as the follower arm). \\
Cameras & Two USB cameras read through OpenCV at 640 $\times$ 480 pixels and 30 frames per second: \texttt{front}, fixed and facing the workspace, and \texttt{wrist}, mounted on the gripper. Note that an overhead camera was specifically not used as subsequent research will be on a mobile platform where this would not be possible. \\
Evaluation computer & Linux workstation with an NVIDIA RTX 5070 Ti (16 GB), 64GB ram, Intel Core Ultra 7 270K CPU. All policy inference ran locally on this machine. \\
Training hardware & Hugging Face Jobs, \texttt{lerobot-gpu} container pinned by digest: A10G for ACT and Diffusion Policy, A100 for SmolVLA and $\pi_{0.5}$ \\
Software & LeRobot at commit \texttt{8352ca62} (21 Sep 2026, development version 0.6.2), with the \texttt{peft} extra added for $\pi_{0.5}$ \\
Workspace & A table-top scene with a white silicone mat. The objects comprise one green felt target ball and two red distractor felt balls. The balls were observed for wear and replaced at regular intervals. \\
Lighting & Illuminance was logged for every block at six points: front camera, gripper camera, workspace, left, right and back using a MASTECH MS8229. \\
\bottomrule
\end{tabular}
\end{table}

Two lighting conditions are used in this paper. \emph{Dim} is the room's ordinary state during testing, 117 to 394 lux at the front camera across the main-comparison blocks. \emph{Bright} is 508 to 725 lux, set up to match the level at which the demonstrations were recorded (about 500 lux). All main-comparison blocks were run in dim light. That choice was planned as a test of robustness; its consequences are examined in Section~\ref{sec:4.3}.

\subsection{Task and placements}\label{sec:3.2}
The task is to pick up a green ball and place it in a basket while two red balls lie on the table as distractors. SmolVLA and $\pi_{0.5}$ are conditioned on language and received the same instruction in training and at test: ``Pick up the green ball and place it in the basket, ignoring the two red distractor balls''. ACT and Diffusion Policy take no language input.

An episode is a success if the green ball is in the basket within 30 seconds of the start. Anything else is a failure, including a ball still held in the gripper when the time runs out. The operator ended an episode as soon as the ball landed in the basket.

Testing used ten fixed placements, labelled N00 to N09. Each placement fixes the positions of the green ball and both red balls, and every model was tested on the same ten. The 10 placements were generated using a pseudo-random number generator.  The first ten ACT blocks were run on twenty placements, N00 to N19, before the protocol was shortened; all comparisons between policies use N00 to N09 only. The N00 to N19 blocks in the ACT runs represented exactly the N00 to N09 sub sets. 

\subsection{Demonstrations, subsets and trimming}\label{sec:3.3}
One operator recorded 225 demonstrations by leader-follower teleoperation at 30 frames per second, in nine batches of 25. Each batch was recorded under slightly different lighting, and the ball positions in each demonstration were drawn pseudorandomly. A demonstration lasts 17 seconds on average (514 frames). 

\paragraph{Subsets.} Twenty-five demonstrations, sampled across all nine batches, were held out and not used for training; they are reserved for a detailed study of validation loss against real success, planned as future research (Section~\ref{sec:8.1}). The other 200 were arranged once into nested subsets of 12, 25, 50, 100 and 200 demonstrations, so that each smaller subset is contained in every larger one. The assignment was fixed with a recorded random seed before any model was trained. Every model at a given budget therefore saw exactly the same demonstrations, and training seeds differ only in training randomness: weight initialisation where applicable, sampling order and noise.

\begin{figure}[t]
\centering
\includegraphics[width=\textwidth]{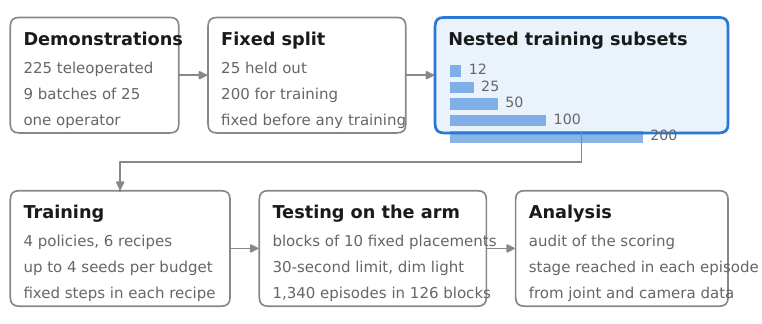}
\caption{\textbf{Study design.} Every model at a given budget is trained on the same demonstrations, and every model is tested on the same ten placements.}
\label{fig:2}
\end{figure}

\paragraph{Trimming.} The demonstrations begin with a stretch in which the operator has not yet moved. After the first Diffusion Policy models failed to start (Section~\ref{sec:4.3}), we made a second copy of the pool with this idle lead-in removed from the start of every demonstration. The cut has a median of 2.8 seconds and removes 17.4\% of all frames. The trimmed copy was used only for the Diffusion Policy models labelled \emph{trimmed}. ACT, SmolVLA and both $\pi_{0.5}$ recipes were trained on the original demonstrations.

\subsection{Policies and training recipes}\label{sec:3.4}
All four policies take the two camera images and the six joint positions as input and output six joint-position targets. Each was trained with LeRobot's implementation and one fixed recipe, applied unchanged at every data budget. Table~\ref{tab:1} lists the recipes; Appendix~\ref{app:F} gives the full configurations.

\begin{table}[t]
\centering
\footnotesize
\caption{\textbf{Training recipes.} Six recipes for four policies. Steps, batch size and learning-rate schedule were the same at every data budget.}
\label{tab:1}
\begin{tabular}{P{2.58cm}P{2.58cm}P{2.58cm}P{1.42cm}P{2.58cm}P{2.58cm}}
\toprule
Recipe & Starting point & Weights trained & Steps $\times$ batch & Learning rate & Actions predicted / executed per chunk \\
\midrule
ACT & From scratch; ResNet-18 image encoder initialised from ImageNet & All & 100,000 $\times$ 8 & 1e-5, constant & 100 / 100 \\
Diffusion Policy, untrimmed demos & From scratch; ResNet-18 image encoder initialised from ImageNet & All & 100,000 $\times$ 64 & 1e-4, cosine decay & 48 / 24 \\
Diffusion Policy, trimmed demos & As above, trained on the trimmed pool & All & 100,000 $\times$ 64 & 1e-4, cosine decay & 48 / 24 \\
SmolVLA & Pretrained \texttt{smolvla\_base} & Action expert; vision-language model frozen & 20,000 $\times$ 64 & 1e-4, cosine decay & 50 / 50 \\
$\pi_{0.5}$, LoRA & Pretrained \texttt{pi05\_base} & Rank-32 low-rank adapters on the action expert: 2.7 million weights, 0.07\% of the model & 6,000 $\times$ 32 & 2.5e-5, decaying to 2.5e-6 & 50 / 50 \\
$\pi_{0.5}$, expert-only & Pretrained \texttt{pi05\_base} & Whole action expert: 693 million of 4.14 billion weights, 16.7\% of the model; vision-language model frozen & 6,000 $\times$ 32 & 2.5e-5, decaying to 2.5e-6 & 50 / 50 \\
\bottomrule
\end{tabular}
\end{table}

The ACT recipe is LeRobot's default and the SmolVLA recipe is its documented fine-tuning command. Diffusion Policy uses the batch size and learning rate of the original real-robot configuration \citep{chi2023diffusion} with a longer prediction horizon. The $\pi_{0.5}$ recipes run 6,000 steps, a length chosen from a LoRA pilot in which training loss was flat from about step 4,000 to step 12,000, and carried over unchanged to the expert-only recipe; the library's examples and the openpi reference configurations run 20,000 to 30,000 steps or more. Section~\ref{sec:5} returns to what these lengths mean for each data budget.

 $\pi_{0.5}$ was fine-tuned twice, with recipes that differ only in which weights are trained (Table~\ref{tab:1b}). We first used low-rank adapters, the least memory-hungry option. After that recipe failed on the robot at every budget, we ran pilots on one data cell (Section~\ref{sec:4.4}) and trained a second grid in which the action expert is trained in full and the vision-language model is frozen, the same division of weights that the SmolVLA recipe uses. Both are reported.

\begin{table}[t]
\centering
\small
\caption{\textbf{The two $\pi_{0.5}$ fine-tuning recipes.} Parameter counts are those printed by the training script.}
\label{tab:1b}
\begin{tabular}{P{4.69cm}P{5.23cm}P{5.23cm}}
\toprule
 & LoRA (first grid) & Expert-only (second grid) \\
\midrule
Base model & \texttt{lerobot/pi05\_base}, 4,143,404,816 weights: a PaliGemma vision-language model and a smaller action expert & Same \\
What is trained & Rank-32 adapters (alpha 32) on the query and value attention projections in the 18 layers of the action expert, and on its four small action and time projections & Every weight of the action expert and its projections; no adapters \\
What stays frozen & Every original weight: the whole vision-language model, and in the action expert the feed-forward blocks, the normalisation layers and the other attention projections & The whole vision-language model, both vision encoder and language model \\
Trainable weights & 2,705,408, or 0.065\% of the model & 693,422,112, or 16.7\% of the model \\
Optimiser & AdamW, weight decay 0.01, gradient clipping at 1.0 & Same \\
Learning rate & 2.5e-5 after 1,000 warm-up steps, then cosine decay to 2.5e-6 at step 6,000 & Same \\
Steps $\times$ batch & 6,000 $\times$ 32 & Same \\
Precision and hardware & 32-bit, gradient checkpointing, one 80 GB A100, about 6.3 seconds per step & Same \\
Models trained / tested & 20 / 11 & 15 / 15 \\
\bottomrule
\end{tabular}
\end{table}

The LoRA recipe uses the library's default adapter placement and changes about one weight in 1,500. Inside the action expert it can alter only low-rank updates to two of the four attention projections in each layer (2.5 million weights) and to the small projections that carry actions and time in and out (0.2 million). The feed-forward blocks and normalisation layers, which together hold about 80\% of each layer's weights, cannot change.

The expert-only recipe trains 256 times as many weights. The 693 million break down, from the model definition, into 311 million in the 18 transformer layers, 117 million in the time-conditioned normalisation layers, 2 million in the projections, and 263 million in a vocabulary output layer that the action expert inherits and never uses when predicting actions. That last layer receives no gradient, so about 430 million weights, 10.4\% of the model, are in fact updated.

Both recipes take the two camera images at 224 $\times$ 224 pixels in two of the model's three camera slots, pad joint state and actions to 32 dimensions with quantile normalisation, predict chunks of 50 actions, and use ten flow-matching steps at inference. Checkpoints were saved at 3,000 and 6,000 steps and the last was tested.

The design called for four training seeds (0 to 3) at each budget. It was completed for ACT and SmolVLA. For the other recipes the number of models trained and tested was cut back, in each case for a stated reason:

\begin{itemize}
\item \emph{Diffusion Policy, untrimmed:} all 20 models were trained; 10 were tested, two seeds per budget, and testing stopped there because the arm did not move.
\item \emph{Diffusion Policy, trimmed:} seeds 1 and 3 were trained and tested at every budget.
\item \emph{$\pi_{0.5}$, LoRA:} three seeds were tested at 12, 25 and 50 demonstrations and one at 100 and 200, and testing stopped after the recipe was judged to have failed.
\item \emph{$\pi_{0.5}$, expert-only:} three models per budget. Seed 0 was trained twice, with and without an exponential moving average (EMA) of the weights, and seed 3 once with EMA. Training jobs for the remaining seeds were terminated by the platform before they finished and could not be rerun in the time available.
\end{itemize}
Models in the main grids were trained without EMA, with the exception just noted. Where a run had EMA enabled we tested the averaged weights. Section~\ref{sec:4.4} shows that EMA made no measurable difference, and on that basis the $\pi_{0.5}$ expert-only blocks with and without EMA are pooled.

We tested the final checkpoint of every run. No checkpoint was selected using robot results or any offline measure.

\subsection{Evaluation protocol}\label{sec:3.5}
The unit of testing is a block: one checkpoint run once on each of the ten placements, in order, with a 30-second limit per episode and a 20-second reset in which the operator re-placed the balls. Illuminance was logged for each block. 

Policies ran through LeRobot's rollout command at a target control rate of 30 Hz, with inference inside the control loop: when the queue of predicted actions empties, the loop waits while the next chunk is computed. $\pi_{0.5}$ was run in 16-bit floating point to fit the 16 GB graphics card. The same settings were used for all policies so that none had a protocol advantage; in particular, asynchronous or real-time chunked execution \citep{black2025rtc} was not used in the main comparison.

Policies were tested in stages on different days (Table~\ref{tab:2}). Within the ACT stage the 20 blocks were run in a shuffled order balanced over seeds and budgets, so that neither drifts with time of day. 

\begin{table}[t]
\centering
\footnotesize
\caption{\textbf{Blocks in the main comparison.} Dim light, standard protocol.}
\label{tab:2}
\begin{tabular}{P{3.64cm}P{1.38cm}P{1.13cm}P{2.01cm}P{2.51cm}P{3.64cm}}
\toprule
Recipe & Episodes & Blocks & Models tested & Test dates (2026) & Front-camera lux, range (median) \\
\midrule
ACT & 300 & 20 & 20 & 23--25 Sep & 223--394 (313) \\
Diffusion Policy, untrimmed demos & 110 & 11 & 10 & 28 Sep--1 Oct & 247--337 (278) \\
SmolVLA & 200 & 20 & 20 & 29 Sep & 135--304 (262) \\
Diffusion Policy, trimmed demos & 150 & 15 & 10 & 1--7 Oct & 207--366 (240) \\
$\pi_{0.5}$, LoRA & 110 & 11 & 11 & 5--6 Oct & 214--362 (272) \\
$\pi_{0.5}$, expert-only & 200 & 20 & 15 & 7--8 Oct & 117--209 (169) \\
\bottomrule
\end{tabular}
\end{table}

The main comparison holds 1,070 episodes in 97 blocks, from 86 trained models. Where a recipe has more blocks than models, the extra blocks are repeat tests of the same checkpoint. Ten ACT blocks hold 20 episodes; all other blocks hold 10.

A further 270 episodes in 29 blocks probe specific questions and are reported where they are used: the same checkpoints in bright light (110 episodes), Diffusion Policy trained with EMA (80), an untrimmed Diffusion Policy control (10), and $\pi_{0.5}$ pilots and probes including real-time chunking and a higher learning rate (70). Appendix~\ref{app:D} lists them.

\subsection{Scoring, audit and automatic measures}\label{sec:3.6}
One operator, who also recorded the demonstrations and was not blind to the model under test, scored each episode as it ran: success or failure, and for failures a short label. Every episode was also recorded in full: measured joint positions and commanded actions at each control step, and video from both cameras.

After testing ended we computed, for every episode, measures that do not depend on the operator. Three of them define a ladder of stages that each episode climbs:

\begin{enumerate}
\item \emph{Started.} An arm joint moved at least 60$^\circ$ from its starting position. The split is clean: no episode classed as not started moved more than 47$^\circ$, and none classed as started moved less than 108$^\circ$.
\item \emph{Reached the ball.} At a gripper close, the green ball was between the jaws in the wrist-camera view. The ball is found by colour thresholding in a fixed region of the image.
\item \emph{Carried the ball.} The ball stayed between the jaws for two seconds or more.
\end{enumerate}
The fourth stage, \emph{delivered}, is the scored success. The stages are cumulative: an episode counts at a stage if it got that far or further, so every delivered episode also counts as carried and reached. Twelve successes in the main comparison had the ball between the jaws for under two seconds. The same recordings give the time of first movement, the achieved control rate, the pause at each new action chunk, and the size of the jump in commanded joint angles from one chunk to the next. Appendix~\ref{app:C} defines each measure.

\paragraph{Outcome classes.} Failures the operator labelled as grasp failures are split by the wrist-camera measure into \emph{location failures}, where the gripper closed but never with the ball between its jaws, and \emph{grip failures}, where the ball was between the jaws at a close and was then lost. Table~\ref{tab:3} gives the classes used in the rest of the paper.

\begin{table}[t]
\centering
\small
\caption{\textbf{Outcome classes over all 1,340 scored episodes.}}
\label{tab:3}
\begin{tabular}{P{5.22cm}P{7.21cm}P{2.73cm}}
\toprule
Outcome & Definition & Episodes \\
\midrule
Success & Green ball delivered to the basket within 30 seconds & 463 \\
Initiation failure & The arm never started & 282 \\
Location failure & The gripper closed one or more times, never with the ball between its jaws & 259 \\
Grip failure & The ball was between the jaws at a close and was then lost & 161 \\
Collision & The arm knocked the ball out of play (79), or collided with the basket (25), the table (19) or the camera (4) & 127 \\
Other & Wrong ball picked (18), still holding at 30 seconds (9), missed the basket (7), dropped in transit (5), started too late (5), no grasp attempt (4) & 48 \\
\bottomrule
\end{tabular}
\end{table}

The wrist-camera measure agrees with the outcomes it should agree with: it found the ball between the jaws in 462 of 463 successes and in none of the 282 episodes in which the arm never started.

We compared the operator's sheet with the recordings for every episode, asking three questions: did the arm move, was the ball ever between the jaws, and was the ball still on the table at the end. The recordings contradicted the sheet in 27 of 1,340 episodes (2.0\%). Twelve doubtful episodes were re-watched on video and ruled on by the operator; the rest were settled by rules that follow from the recordings, for example that an arm which never moved cannot have succeeded.

Sixteen scored successes became failures, six scored failures became successes, and five failures changed class. Ten of the changes were five pairs of neighbouring rows entered one row off or swapped, which alters which placement received which result but not the block total. The number of successes fell from 473 to 463. All results in this paper use the audited scores, and Appendix~\ref{app:B} lists every change.

\subsection{Statistics}\label{sec:3.7}
The unit of analysis is the episode, grouped into blocks. We use the following throughout.

\begin{itemize}
\item \textbf{Success rates} are reported as successes over episodes with 95\% Wilson score intervals \citep{wilson1927probable}.
\item \textbf{Two cells} are compared with Fisher's exact test, and differences in rate carry Newcombe hybrid-score intervals \citep{newcombe1998interval}.
\item \textbf{Trend with data} is a logistic regression of success on log2 of the number of demonstrations, fitted by generalised estimating equations with blocks as clusters \citep{liang1986gee}. We report the odds ratio per doubling of data.
\item \textbf{Spread between blocks} is the Pearson chi-square of block success counts about their cell's pooled rate, divided by its degrees of freedom. A ratio of 1 means blocks differ as much as independent coin flips would; larger values mean seeds or repeat runs disagree by more than chance.
\item \textbf{Seeds within a budget} are compared with an exact test of homogeneity on the seeds-by-outcome table. Whether a seed is consistently better or worse across budgets is tested with a likelihood-ratio test for a seed term in a logistic model that has budget as a factor.
\item \textbf{Paired comparisons} of two conditions on the same placements use the exact McNemar test.
\item \textbf{Repeat tests} of one checkpoint are summarised by the number of placements whose outcome changed, set against the number expected if every episode were an independent draw at the pair's average rate.
\end{itemize}
We do not correct for multiple comparisons, and p-values should be read as descriptive. A cell of 40 episodes has a 95\% interval of about $\pm$15 percentage points, so differences between two cells smaller than about 25 points are not resolved by this study.

\FloatBarrier
\section{Results}\label{sec:4}
Unless stated otherwise, results are from the main comparison: dim light, ten placements, audited scores. For ACT we use its first ten placements so that every recipe is measured on the same ten.

\subsection{Success against number of demonstrations}\label{sec:4.1}
\begin{figure}[t]
\centering
\includegraphics[width=\textwidth]{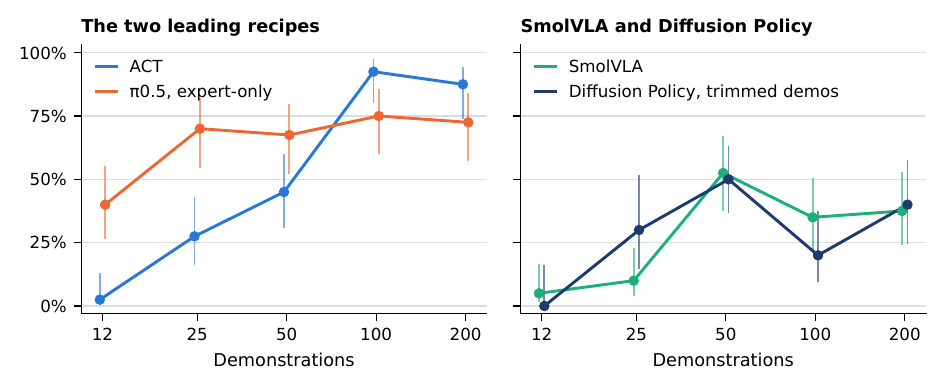}
\caption{\textbf{Success rate against number of demonstrations for the four working recipes.}}
\label{fig:3}
\end{figure}

The four working recipes scale in three different ways (Figure~\ref{fig:3}, Table~\ref{tab:4}).

\begin{table}[t]
\centering
\footnotesize
\caption{\textbf{Success by recipe and number of demonstrations.} Each cell gives successes over episodes and, in brackets, the rate with its 95\% Wilson interval in percent. The last column is the odds ratio per doubling of data, with its 95\% interval.}
\label{tab:4}
\begin{tabular}{P{2.63cm}P{1.81cm}P{1.81cm}P{1.81cm}P{1.81cm}P{1.81cm}P{2.35cm}}
\toprule
Recipe & 12 demos & 25 demos & 50 demos & 100 demos & 200 demos & Odds ratio per doubling \\
\midrule
ACT & 1/40 (3; 0--13) & 11/40 (28; 16--43) & 18/40 (45; 31--60) & 37/40 (93; 80--97) & 35/40 (88; 74--95) & 3.89 (2.70--5.62) \\
$\pi_{0.5}$, expert-only & 16/40 (40; 26--55) & 28/40 (70; 55--82) & 27/40 (68; 52--80) & 30/40 (75; 60--86) & 29/40 (73; 57--84) & 1.37 (1.15--1.63) \\
SmolVLA & 2/40 (5; 1--17) & 4/40 (10; 4--23) & 21/40 (53; 37--67) & 14/40 (35; 22--50) & 15/40 (38; 24--53) & 1.61 (1.21--2.14) \\
Diffusion Policy, trimmed demos & 0/20 (0; 0--16) & 6/20 (30; 15--52) & 25/50 (50; 37--63) & 6/30 (20; 10--37) & 12/30 (40; 25--58) & 1.32 (0.88--1.97) \\
Diffusion Policy, untrimmed demos & 0/20 (0; 0--16) & 1/20 (5; 1--24) & 0/20 (0; 0--16) & 0/30 (0; 0--11) & 2/20 (10; 3--30) & 1.79 (0.48--6.70) \\
$\pi_{0.5}$, LoRA & 1/30 (3; 1--17) & 2/30 (7; 2--21) & 1/30 (3; 1--17) & 0/10 (0; 0--28) & 0/10 (0; 0--28) & 0.71 (0.40--1.28) \\
\bottomrule
\end{tabular}
\end{table}

\paragraph{ACT follows a steep S-curve.} It succeeds in 1 of 40 episodes at 12 demonstrations, 18 of 40 at 50 and 37 of 40 at 100, and is unchanged at 200 (35 of 40; p = 0.71 for the last step). Each doubling of data multiplies its odds of success by 3.9. Over all 20 placements, where they were run, the counts are 2, 18, 26, 51 and 53 of 60, the same shape.

\paragraph{$\pi_{0.5}$ with expert-only fine-tuning starts high and saturates early.} It succeeds in 16 of 40 episodes at 12 demonstrations and 28 of 40 at 25 (p = 0.013), and then does not move: 27, 30 and 29 of 40 at 50, 100 and 200.

\paragraph{SmolVLA and Diffusion Policy peak at 50 demonstrations and fall back.} SmolVLA jumps from 4 of 40 at 25 demonstrations to 21 of 40 at 50 (p < 0.001), then drops to 14 and 15 of 40. Diffusion Policy on trimmed demonstrations reaches 25 of 50 at 50 demonstrations and falls to 6 of 30 at 100 (p = 0.009) before a partial recovery to 12 of 30. Sections~\ref{sec:4.3} and~\ref{sec:4.5} examine both falls: SmolVLA's is a change in how often the arm starts, and Diffusion Policy's comes mostly from one seed that scored 0 of 20.

\paragraph{The order of the two leading recipes reverses between 50 and 100 demonstrations.} $\pi_{0.5}$ leads ACT by 38 percentage points at 12 demonstrations and 43 at 25 (both p < 0.001), and by 23 at 50 (p = 0.07). ACT leads by 18 points at 100 (p = 0.07) and 15 at 200 (p = 0.16). The direction reverses, but with 40 episodes per cell the gaps at 50 and 100 demonstrations are borderline and the gap at 200 is not resolved. The two recipes were also tested on different days in different light (Section~\ref{sec:7}). What the data establish is that the pretrained policy has a large advantage at 12 and 25 demonstrations and no measurable advantage at 100 or 200.

\paragraph{Two recipes barely worked.} Diffusion Policy trained on the untrimmed demonstrations succeeded in 3 of 110 episodes, and $\pi_{0.5}$ with LoRA in 4 of 110, with no trend in either. Sections~\ref{sec:4.3} and~\ref{sec:4.4} take them up.

\subsection{Where episodes fail}\label{sec:4.2}
A success rate says how often a policy finished the task. The recordings also say how far it got when it did not. This subsection uses all main-comparison episodes, including all 60 per budget for ACT.

\begin{figure}[t]
\centering
\includegraphics[width=\textwidth]{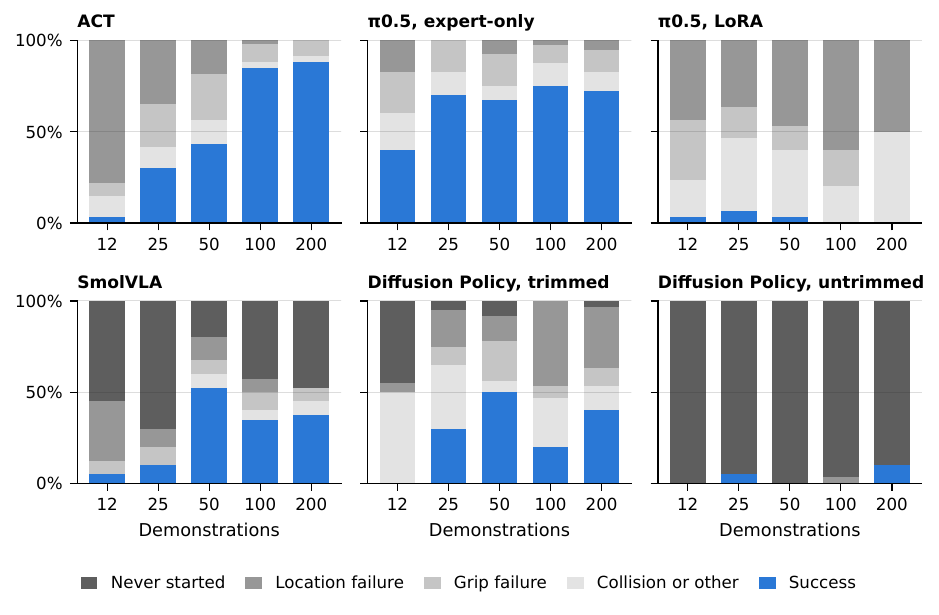}
\caption{\textbf{Outcome of every episode, by recipe and number of demonstrations.}}
\label{fig:4}
\end{figure}

\begin{table}[t]
\centering
\footnotesize
\caption{\textbf{Episodes reaching each stage of the task, all budgets pooled.}}
\label{tab:5}
\begin{tabular}{P{4.28cm}P{1.62cm}P{1.48cm}P{2.80cm}P{1.92cm}P{2.21cm}}
\toprule
Recipe & Episodes & Started & Reached the ball & Carried it & Delivered it \\
\midrule
ACT & 300 & 300 & 208 & 157 & 150 \\
$\pi_{0.5}$, expert-only & 200 & 200 & 177 & 135 & 130 \\
SmolVLA & 200 & 106 & 75 & 58 & 56 \\
Diffusion Policy, trimmed demos & 150 & 135 & 80 & 55 & 49 \\
Diffusion Policy, untrimmed demos & 110 & 4 & 3 & 3 & 3 \\
$\pi_{0.5}$, LoRA & 110 & 110 & 39 & 13 & 4 \\
\bottomrule
\end{tabular}
\end{table}

\paragraph{The working recipes differ early in the task and hardly at all late in it.} Among the four working recipes, the share of started episodes that reach the ball ranges from 59\% (Diffusion Policy) to 89\% ($\pi_{0.5}$). After that the recipes look alike: between 69\% and 77\% of reaches become a carry, and between 89\% and 97\% of carries are delivered. What separates the policies is whether the arm starts and whether the gripper arrives on the ball.

\paragraph{More data moves ACT's failures later in the task, then removes them.} At 12 demonstrations ACT always moves but closes its gripper off the ball in 47 of 60 episodes. At 25 and 50 it reaches the ball in most episodes (36 and 47 of 60) and then loses it, with 14 and 15 grip failures. At 100 and 200 it reaches the ball in 59 and 60 of 60 episodes and delivers nearly every ball it carries.

\paragraph{$\pi_{0.5}$'s plateau is a grasping problem, not a reaching problem.} With expert-only fine-tuning it reaches the ball in 37 or 38 of 40 episodes at every budget from 25 demonstrations upward, but carries it in only 28 to 31. Of the 160 episodes at those budgets, 23 ended in a grip failure and 12 with the ball knocked away, against 6 location failures. The LoRA recipe fails earlier: it moves in every episode but puts the gripper on the ball in only 39 of 110.

\paragraph{Reaching accuracy improves with data even where success does not.} Table~\ref{tab:6} isolates the first grasp attempt of each episode, before any recovery. For ACT, SmolVLA and $\pi_{0.5}$ expert-only the share of first closes that land on the ball rises with data more steadily than success rate does, with only small reversals. The exception is Diffusion Policy at 100 demonstrations, where the first close was on the ball in 3 of 30 episodes and the median miss was 120 pixels in the wrist view, against 28 pixels at 50 demonstrations.

\begin{table}[t]
\centering
\footnotesize
\caption{\textbf{First grasp attempt: closes with the ball between the jaws, over episodes in which the gripper closed.} In brackets, the median distance in pixels between the ball and the jaw centre at that close, in a 320-pixel-wide wrist view.}
\label{tab:6}
\begin{tabular}{P{4.42cm}P{1.98cm}P{1.98cm}P{1.98cm}P{1.98cm}P{1.98cm}}
\toprule
Recipe & 12 demos & 25 demos & 50 demos & 100 demos & 200 demos \\
\midrule
ACT & 4/60 (74) & 24/60 (51) & 33/60 (45) & 53/60 (18) & 51/60 (27) \\
$\pi_{0.5}$, expert-only & 24/40 (38) & 33/40 (23) & 31/40 (25) & 37/40 (19) & 34/40 (26) \\
SmolVLA & 3/18 (75) & 7/12 (37) & 21/31 (24) & 17/22 (28) & 17/20 (14) \\
Diffusion Policy, trimmed demos & 3/11 (108) & 12/19 (38) & 36/46 (28) & 3/30 (120) & 12/27 (46) \\
$\pi_{0.5}$, LoRA & 1/30 (92) & 7/28 (79) & 3/30 (135) & 1/10 (83) & 0/9 (119) \\
\bottomrule
\end{tabular}
\end{table}

\subsection{Arms that never start}\label{sec:4.3}
In 282 of the 1,340 episodes the arm made no real movement. The servos held position and the joints typically drifted by a few degrees (a median of 3$^\circ$; 32 episodes moved more than 10$^\circ$ and none more than 47$^\circ$), but no attempt on the task began. In these episodes the commanded actions themselves stayed near the start pose (a median excursion of 6$^\circ$ for Diffusion Policy on untrimmed demonstrations and 7$^\circ$ for SmolVLA), so the arm was doing what the policy told it to do.

\begin{table}[t]
\centering
\footnotesize
\caption{\textbf{Episodes in which the arm never started, over episodes run.}}
\label{tab:7}
\begin{tabular}{P{4.24cm}P{1.61cm}P{1.61cm}P{1.61cm}P{1.75cm}P{1.75cm}P{1.46cm}}
\toprule
Recipe & 12 demos & 25 demos & 50 demos & 100 demos & 200 demos & All \\
\midrule
ACT & 0/60 & 0/60 & 0/60 & 0/60 & 0/60 & 0/300 \\
$\pi_{0.5}$, expert-only & 0/40 & 0/40 & 0/40 & 0/40 & 0/40 & 0/200 \\
$\pi_{0.5}$, LoRA & 0/30 & 0/30 & 0/30 & 0/10 & 0/10 & 0/110 \\
SmolVLA & 22/40 & 28/40 & 8/40 & 17/40 & 19/40 & 94/200 \\
Diffusion Policy, untrimmed demos & 20/20 & 19/20 & 20/20 & 29/30 & 18/20 & 106/110 \\
Diffusion Policy, trimmed demos & 9/20 & 1/20 & 4/50 & 0/30 & 1/30 & 15/150 \\
\bottomrule
\end{tabular}
\end{table}

\paragraph{Trimming the idle lead-in removed the failure in Diffusion Policy.} Trained on the original demonstrations, Diffusion Policy never started in 106 of 110 episodes. Trained by the same recipe on the same demonstrations with the idle lead-in removed, it failed to start in 15 of 150, nine of those at 12 demonstrations. An untrimmed model retrained as a control and tested alongside the trimmed models failed to start in 9 of 10 episodes, which rules out a change in the rig between the two periods. The simplest reading is that the policy learned, from the opening frames of each demonstration, that the correct action at the start pose is to stay there. Two cautions apply. The trimmed and untrimmed models were tested on different days and are not the same seeds at every budget. And untrimmed data does not always produce the failure: Section~\ref{sec:4.4} reports two retrained untrimmed models that started in most episodes.

\begin{figure}[t]
\centering
\includegraphics[width=\textwidth]{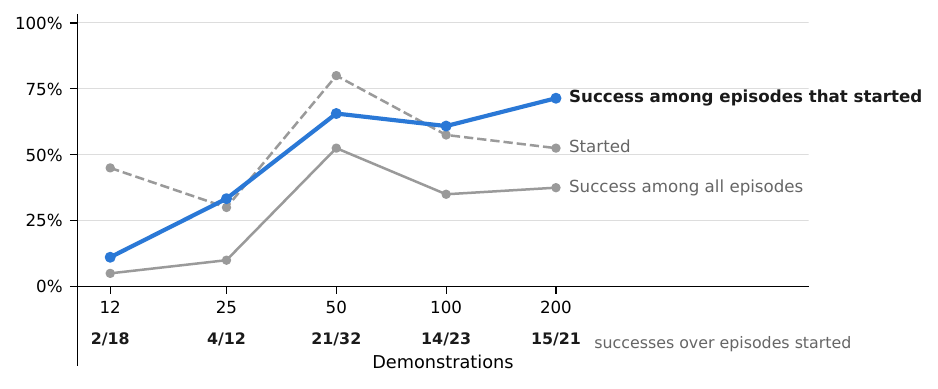}
\caption{\textbf{SmolVLA by number of demonstrations: episodes that started, and success counted over started episodes and over all episodes.}}
\label{fig:5}
\end{figure}

\paragraph{SmolVLA has the same symptom, and it explains the shape of its curve.} SmolVLA was trained on the untrimmed demonstrations and never started in 94 of 200 episodes. It froze least at 50 demonstrations (8 of 40), which is where its raw success peaks. Counting only episodes that started, success rises to 50 demonstrations and then holds: 2 of 18, 4 of 12, 21 of 32, 14 of 23 and 15 of 21, or 11\%, 33\%, 66\%, 61\% and 71\% (Figure~\ref{fig:5}). The dip in Section~\ref{sec:4.1} is a change in how often the arm starts, not in what it does once it has.

\paragraph{When SmolVLA starts, it starts late.} Among episodes that started, the median time to the first 20$^\circ$ of movement is 1.1 seconds for ACT and for Diffusion Policy on trimmed demonstrations, 1.3 seconds for $\pi_{0.5}$ expert-only, and 3.6 seconds for SmolVLA, with one SmolVLA start in ten coming after 16 seconds. Late starts leave little of the 30-second limit. ACT shows a milder form of the same behaviour: on the ten shared placements its median start moves from 0.6 seconds at 12 demonstrations to 1.3 seconds at 200, consistent with reproducing more of the lead-in the more demonstrations it has seen.

\paragraph{Brighter light helped Diffusion Policy start, but far less than trimming did.} Five untrimmed Diffusion Policy checkpoints and five SmolVLA checkpoints were run for ten episodes each in dim and in bright light. Diffusion Policy started in 16 of 50 bright episodes against 2 of 50 dim ones, and its successes went from 2 to 8. SmolVLA started in 36 of 50 bright episodes against 32 of 50 dim, and succeeded in 10 against 15. The gap in illuminance between recording and testing therefore contributed to Diffusion Policy's freezing but does not explain SmolVLA's.

\paragraph{ACT and $\pi_{0.5}$ never froze.} Both were trained on the same untrimmed demonstrations and started in every one of 610 episodes across three recipes. Section~\ref{sec:6} discusses why the same data produces the failure in two policies and not in the others.

\subsection{Fine-tuning recipe and EMA}\label{sec:4.4}
Two changes to how a policy was trained had effects as large as most changes in the amount of data. Table~\ref{tab:8} sets them beside the changes that had no measurable effect.

\begin{table}[t]
\centering
\footnotesize
\caption{\textbf{Effect of recipe and condition changes on success.} Differences are in percentage points with 95\% intervals that treat episodes as independent, so they understate the uncertainty where blocks disagree. The two rows marked paired compare the same placements and use an interval for paired differences.}
\label{tab:8}
\begin{tabular}{P{4.03cm}P{4.03cm}P{2.50cm}P{1.39cm}P{2.64cm}}
\toprule
Change & Compared & With the change & Without & Difference \\
\midrule
Trimming the idle lead-in (Diffusion Policy) & All budgets & 49/150 & 3/110 & +30 (21 to 38) \\
Expert-only instead of LoRA ($\pi_{0.5}$) & All budgets & 130/200 & 4/110 & +61 (53 to 68) \\
EMA weights ($\pi_{0.5}$ expert-only) & Seed 0, five budgets, same day, same placements (paired) & 33/50 & 34/50 & $-$2 ($-$19 to 15) \\
EMA enabled in training (Diffusion Policy, trimmed) & Same four cells & 15/40 & 31/80 & $-$1 ($-$18 to 17) \\
Averaged instead of final weights (Diffusion Policy, same run) & Two checkpoints (paired) & 6/20 & 8/20 & $-$10 ($-$37 to 17) \\
Bright light (Diffusion Policy, untrimmed) & Five checkpoints & 8/50 & 2/50 & +12 ($-$0.2 to 25) \\
Bright light (SmolVLA) & Five checkpoints & 10/50 & 15/50 & $-$10 ($-$26 to 7) \\
Ten times the learning rate ($\pi_{0.5}$ LoRA) & One cell, 100 demos & 5/10 & 0/10 & +50 (12 to 76) \\
Real-time chunking ($\pi_{0.5}$ LoRA) & Three checkpoints & 1/30 & 0/30 & +3 \\
\bottomrule
\end{tabular}
\end{table}

\paragraph{For $\pi_{0.5}$, which weights are trained decided the outcome.} With rank-32 low-rank adapters, which train 2.7 million weights or 0.07\% of the model, $\pi_{0.5}$ succeeded in 4 of 110 episodes and showed no trend with data. With the action expert trained in full and the vision-language model frozen, 693 million weights or 16.7\% of the model, it succeeded in 130 of 200. The data, the base model, the optimiser, the number of steps, the batch size and the learning-rate schedule were the same (Table~\ref{tab:1b}). The LoRA models were not inert: the arm moved in every episode and headed for the scene, but the gripper arrived on the ball in only 39 of 110 episodes (Table~\ref{tab:5}).

\paragraph{How the working recipe was found.} After the adapter grid had failed, four variants were trained on one cell, 100 demonstrations with seed 0, each changing one thing (Table~\ref{tab:8b}). Raising the adapter learning rate tenfold took that cell from 0 to 5 of 10. Training the whole action expert at the original learning rate scored 9 of 10. The expert-only recipe was then adopted without further change. A 200-demonstration model was trained and tested first as a gate, and scored 8 of 10, before the rest of the grid was launched.

\begin{table}[t]
\centering
\footnotesize
\caption{\textbf{$\pi_{0.5}$ pilots on one cell: 100 demonstrations, seed 0, ten episodes.} All variants use 6,000 steps at batch 32.}
\label{tab:8b}
\begin{tabular}{P{4.69cm}P{3.23cm}P{3.40cm}P{3.56cm}}
\toprule
Variant & Trainable weights & Peak learning rate & Successes \\
\midrule
LoRA, standard recipe (the grid model) & 2,705,408 & 2.5e-5 & 0/10 \\
LoRA, ten times the learning rate & 2,705,408 & 2.5e-4 & 5/10 \\
Wide LoRA: adapters on the attention and feed-forward layers of both the language model and the action expert & 53,282,816 & 2.5e-5 & trained, not tested \\
Wide LoRA, ten times the learning rate & 53,282,816 & 2.5e-4 & trained, not tested \\
Expert-only & 693,422,112 & 2.5e-5 & 9/10 \\
\bottomrule
\end{tabular}
\end{table}

\paragraph{The pilots suggest the LoRA recipe was under-fitted, not incapable.} The tenfold learning rate is the only evidence, from one model and ten episodes. Training the standard adapters for longer did not help in two five-episode probes: checkpoints at 12,000 and 15,000 steps scored 0 of 5 and 1 of 5. Running three LoRA checkpoints with real-time chunking instead of in-loop inference did not help either (1 of 30 against 0 of 30), so the failure lay in what the model had learned and not in how its actions were executed. We did not pursue a tuned LoRA recipe. The result is a warning that an adapter recipe that trains without error can still fall far short at a learning rate and length that suit the action expert.

\paragraph{A caution on how the recipe was chosen.} The expert-only recipe was selected on the strength of ten robot episodes in one cell, so its choice is not independent of the test bench. The pilot model is not part of the main comparison. The two grid models for the same cell were trained separately (the one without EMA scored 6 and then 8 of 10), and the other 170 main-comparison episodes come from cells the pilot never touched.

\paragraph{EMA made no measurable difference.} For $\pi_{0.5}$ expert-only, the seed-0 models trained with and without EMA were tested on the same day on the same placements: 33 of 50 with EMA against 34 of 50 without, with 9 placements won only by the EMA model and 10 only by the other (exact McNemar p = 1.0). For Diffusion Policy on trimmed demonstrations, runs with EMA enabled scored 15 of 40 against 31 of 80 for runs without, and the averaged weights of a run scored 6 of 20 against 8 of 20 for its final weights. In these recipes the learning rate decays to a small fraction of its starting value, which leaves the averaged and final weights close together.

\paragraph{One observation we report without interpreting.} Two untrimmed Diffusion Policy models retrained with EMA enabled failed to start in 8 of 20 episodes, against 27 of 30 for the original runs of the same two cells (p < 0.001). The blocks were run on different days, and success barely changed (5 of 20 against 3 of 30). Two checkpoints are too few to separate an effect of EMA from run-to-run variation.

\subsection{Seed and re-test variability}\label{sec:4.5}
Two sources of variation sit under every cell in Table~\ref{tab:4}. The same checkpoint does not score the same twice, and different training seeds give different checkpoints. This subsection measures each and asks, policy by policy, whether seeds differ by more than the measurement can explain.

A seed here means the same demonstrations and recipe with different training randomness. For the two policies trained from scratch, the seed also fixes the initial weights, which are therefore shared by, say, every seed-0 model across the five budgets. For the two pretrained policies the seed affects only sampling order and noise.

\begin{figure}[t]
\centering
\includegraphics[width=\textwidth]{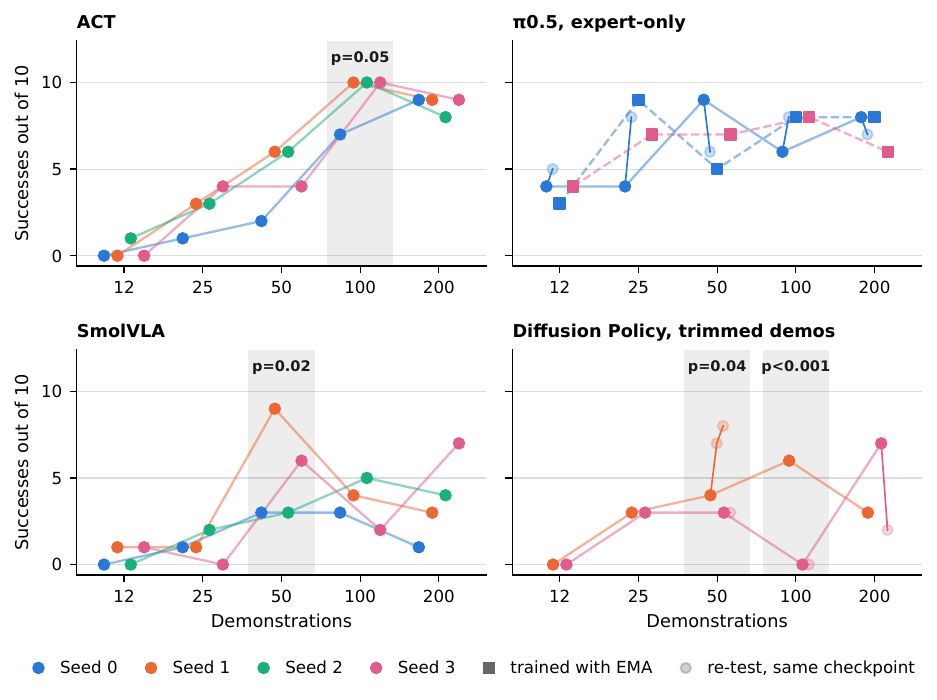}
\caption{\textbf{Every block of the four working recipes, by seed.} Colour gives the training seed, and within a budget the models also run left to right in seed order. For $\pi_{0.5}$ the three models are seed 0 without EMA (circles, tested twice), seed 0 with EMA and seed 3 with EMA (squares). Diffusion Policy has seeds 1 and 3 only.}
\label{fig:6}
\end{figure}

\paragraph{Re-testing the same checkpoint.} Twelve pairs of blocks ran the same checkpoint twice on the same ten placements. The two counts differed by 2.0 successes out of 10 on average, and 40 of the 120 placements changed outcome. If every episode were an independent draw at the pair's average rate, 44 changes would be expected. A result on a single placement is therefore only weakly repeatable. Placement does carry some signal: in the ten pairs with any success, a success was followed by a success in 39 of 56 cases and a failure by a success in 23 of 44, and Appendix~\ref{app:E} shows placement effects. But one block cannot tell a mastered placement from a lucky one. One consequence is that two blocks of ten must differ by about six successes before the difference is distinguishable from chance (2 against 8 gives p = 0.02; 1 against 6 gives p = 0.06).

\begin{table}[t]
\centering
\scriptsize
\caption{\textbf{Do seeds differ?} For each budget, the p-value of an exact test that all models at that budget share one success rate. The last two columns give the spread between blocks as a ratio to what independent draws would produce (1 means no excess), first between different models and then between repeat tests of the same checkpoint.}
\label{tab:9}
\begin{tabular}{P{2.50cm}P{1.73cm}P{0.95cm}P{0.95cm}P{0.95cm}P{1.04cm}P{1.04cm}P{2.07cm}P{2.24cm}}
\toprule
Recipe & Models per budget & 12 demos & 25 demos & 50 demos & 100 demos & 200 demos & Spread between models & Spread between re-tests \\
\midrule
ACT & 4 & 1.00 & 0.60 & 0.27 & 0.049 & 1.00 & 1.4 (p = 0.16) & not re-tested \\
$\pi_{0.5}$, expert-only & 3 & 0.91 & 0.27 & 0.45 & 0.80 & 0.72 & 1.3 (p = 0.25) & 1.4 (p = 0.21) \\
SmolVLA & 4 & 1.00 & 0.89 & 0.022 & 0.69 & 0.057 & 1.6 (p = 0.06) & not re-tested \\
Diffusion Policy, trimmed demos & 2 & 1.00 & 1.00 & 0.042 & < 0.001 & 0.69 & 3.0 (p = 0.02) & 2.2 (p = 0.07) \\
\bottomrule
\end{tabular}
\end{table}

\paragraph{ACT: seeds behave as repeat draws, with a hint of a weaker initialisation.} Within a budget the four seeds differ no more than chance allows. The one nominal exception is 100 demonstrations, where seed 0 scored 7 of 10 and the other three scored 10; among the 20 tests in Table~\ref{tab:9}, one result at that level is expected by chance. Seed 0 is also the lowest or equal lowest at four of five budgets, 19 of 50 overall against 27 or 28 for each other seed. The test for a seed term across budgets gives p = 0.03 on the ten shared placements and p = 0.07 over all placements. Because the seed fixes ACT's initial weights, a persistent effect of initialisation is possible, but seed 0 was singled out after seeing the data and the evidence is weak. We do not claim it.

\paragraph{$\pi_{0.5}$: no detectable difference between models.} The three expert-only models at each budget are statistically indistinguishable at every budget, and seed 0 and seed 3 score 98 of 150 and 32 of 50 overall. Its repeat tests vary as independent draws would.

\paragraph{SmolVLA: seeds differ, and what differs is whether the arm starts.} At 50 demonstrations the four seeds scored 3, 9, 3 and 6 of 10, and at 200 they scored 1, 3, 4 and 7. At 200 demonstrations the seeds failed to start in 8, 7, 4 and 0 of 10 episodes (p = 0.001), and among episodes that started their success is alike: 1 of 2, 3 of 3, 4 of 6 and 7 of 10 (p = 0.75). At 50 the pattern is similar but less clean: 1, 0, 5 and 2 failures to start (p = 0.05), with success among started episodes of 3 of 9, 9 of 10, 3 of 5 and 6 of 8 (p = 0.07). No seed is consistently better across budgets (p = 0.07 for success, p = 0.10 for starting). Training randomness in SmolVLA appears mainly to set how strongly a given model has absorbed the idle lead-in.

\paragraph{Diffusion Policy: the least repeatable policy, between seeds and between runs.} Blocks vary three times more than independent draws would. At 100 demonstrations seed 1 scored 6 of 10 and seed 3 scored 0 of 10, then 0 of 10 again on re-test. Both seeds moved in every episode and both missed with their first grasp (first close on the ball in 2 of 10 and 1 of 20 episodes), but seed 1 recovered and reached the ball in 7 of 10 episodes while seed 3 did so in 2 of 20. At 200 demonstrations the order reverses: seed 1 scored 3 of 10 and seed 3 scored 7, then 2 on re-test six days later. With two seeds per budget, the dip at 100 demonstrations in Figure~\ref{fig:3} cannot be separated from seed and run variability.

\paragraph{Seeds differ less in reaching than in outcome.} The accuracy of the first grasp attempt (Table~\ref{tab:6}) differs between seeds in only one cell out of 20, ACT at 25 demonstrations. Seed-to-seed differences in success come mainly from starting (SmolVLA) and from recovery after a missed grasp (Diffusion Policy), and not from how well the arm is aimed.

\paragraph{What this design can and cannot resolve.} A single seed tested on ten episodes carries an interval of about $\pm$25 percentage points, so per-seed numbers should not be read individually. Four seeds of ten episodes detect seed effects only when they are as large as SmolVLA's 3 against 9. For ACT and $\pi_{0.5}$ the study shows that any seed effect is small relative to that resolution; it does not show that there is none.

\subsection{Control behaviour}\label{sec:4.6}
The recordings also show how each policy drives the arm, independent of whether the episode succeeded. Table~\ref{tab:10} summarises episodes in which the arm started.

\begin{table}[t]
\centering
\footnotesize
\caption{\textbf{Control and timing, episodes in which the arm started.} Medians. The pause is the stall at each new action chunk, estimated from the frames recorded in full-length episodes. The chunk jump is the step in commanded joint angles where one chunk ends and the next begins, measured during the reach.}
\label{tab:10}
\begin{tabular}{P{2.18cm}P{1.50cm}P{1.50cm}P{1.65cm}P{1.28cm}P{1.50cm}P{1.95cm}P{2.18cm}}
\toprule
Recipe & Actions per chunk & Control rate (Hz) & Pause per chunk (s) & Chunk jump ($^\circ$) & Time to start (s) & Length of a success (s) & Successes whose first close was on the ball \\
\midrule
ACT & 100 & 29.9 & 0.01 & 3.1 & 1.1 & 16.5 & 127/150 \\
$\pi_{0.5}$, expert-only & 50 & 27.3 & 0.17 & 1.9 & 1.3 & 16.8 & 121/130 \\
SmolVLA & 50 & 28.7 & 0.08 & 1.5 & 3.6 & 22.0 & 50/56 \\
Diffusion Policy, trimmed demos & 24 & 28.7 & 0.04 & 2.6 & 1.1 & 17.1 & 41/49 \\
$\pi_{0.5}$, LoRA & 50 & 27.0 & 0.19 & 4.5 & 3.0 & n/a & n/a \\
\bottomrule
\end{tabular}
\end{table}

\paragraph{All four policies hold close to the 30 Hz target on a 16 GB consumer graphics card.} Inference ran inside the control loop, so each new chunk stalls the arm briefly: 0.01 seconds for ACT and 0.17 for $\pi_{0.5}$, which over an episode costs $\pi_{0.5}$ about a tenth of its control steps. We did not test whether these stalls affect success.

\paragraph{Agreement between successive chunks improves with data only for the policies trained from scratch.} The jump in commanded joint angles at a chunk boundary falls steadily with data for ACT (4.6$^\circ$ at 12 demonstrations to 2.3$^\circ$ at 200) and for Diffusion Policy (8.1$^\circ$ to 1.0$^\circ$). For the pretrained policies it does not change: SmolVLA and $\pi_{0.5}$ expert-only stay between 1.4$^\circ$ and 2.1$^\circ$ at every budget. $\pi_{0.5}$ with LoRA stays between 4.0$^\circ$ and 5.0$^\circ$, so this continuous measure separates the working $\pi_{0.5}$ recipe from the failed one at every budget, including those where neither could be told apart by success at ten episodes.

\paragraph{SmolVLA is slower to finish.} Successful episodes last a median of 16.5 to 17.1 seconds for ACT, $\pi_{0.5}$ and Diffusion Policy, and 22.0 seconds for SmolVLA, mostly because of its late starts. The 30-second limit is not neutral. Nine main-comparison episodes ended with the ball still in the gripper (three for ACT, three for $\pi_{0.5}$ LoRA, two for SmolVLA and one for Diffusion Policy), and three SmolVLA episodes started too late to finish.

\paragraph{Most successes came from a first grasp that was on target.} The first close was on the ball in 93\% of the successes of $\pi_{0.5}$, 89\% of those of SmolVLA, 85\% of those of ACT and 84\% of those of Diffusion Policy. The rest succeeded on a later attempt.

\FloatBarrier
\section{Training exposure: the fixed-step confound}\label{sec:5}
Every recipe trained for a fixed number of steps at a fixed batch size, whatever the data budget. The number of training samples drawn was therefore constant within a recipe, and the number of times each frame was seen fell in proportion as the data grew. This section states what that does to the interpretation of Section~\ref{sec:4}.

\subsection{How many times each model saw its data}\label{sec:5.1}
We count exposure in passes: one pass is one training sample drawn for every frame in the training set. The 200-demonstration set holds 102,792 frames, and trimming removes 17.4\% of them.

Table~\ref{tab:11} gives the number of passes for every recipe and budget.

\begin{table}[t]
\centering
\footnotesize
\caption{\textbf{Passes over the training data.} Samples drawn are steps times batch size.}
\label{tab:11}
\begin{tabular}{P{3.99cm}P{2.20cm}P{1.51cm}P{1.51cm}P{1.51cm}P{1.65cm}P{1.65cm}}
\toprule
Recipe & Samples drawn & 12 demos & 25 demos & 50 demos & 100 demos & 200 demos \\
\midrule
Diffusion Policy, trimmed demos & 6.4 million & 1,256 & 603 & 302 & 151 & 75 \\
Diffusion Policy, untrimmed demos & 6.4 million & 1,038 & 498 & 249 & 125 & 62 \\
SmolVLA & 1.28 million & 208 & 100 & 50 & 25 & 12 \\
ACT & 0.8 million & 130 & 62 & 31 & 16 & 7.8 \\
$\pi_{0.5}$, both recipes & 0.19 million & 31 & 15 & 7.5 & 3.7 & 1.9 \\
\bottomrule
\end{tabular}
\end{table}

Two things follow. First, the recipes are not comparable in exposure: at any one budget Diffusion Policy saw its data 33 to 40 times more often than $\pi_{0.5}$. Second, within a recipe the data budget and the exposure are tied together exactly. Each doubling of data halves the passes, so no curve in this study can by itself separate the effect of more data from the effect of less training on each example.

\subsection{What the robot data say about over- and under-training}\label{sec:5.2}
The behavioural measures give circumstantial evidence, recipe by recipe.

\begin{itemize}
\item \textbf{ACT (130 to 7.8 passes).} There is no sign that the many passes at small budgets did harm: those models fail by missing the ball, which is what too little data looks like. At 200 demonstrations each frame was seen about eight times. Success is flat from 100 to 200 (37 and 35 of 40) and first-grasp accuracy is slightly worse at 200 (Table~\ref{tab:6}), while chunk-to-chunk agreement keeps improving. The flat step could be a ceiling near 90\% or mild under-training.
\item \textbf{$\pi_{0.5}$, expert-only (31 to 1.9 passes).} This is the strongest candidate for under-training. Nothing we measured improves from 25 to 200 demonstrations, the 200-demonstration models saw each frame about twice, and the run is a fifth to a third as long as published fine-tuning configurations. The schedule length was set from the loss curve of the LoRA recipe and was not re-examined for this one. A ceiling set by grasp execution would produce the same flat curve.
\item \textbf{SmolVLA (208 to 12 passes).} Among episodes that start, success rises to 50 demonstrations and holds to 200, and first-grasp accuracy keeps rising. The manipulation skill shows no sign of decline at the larger budgets; the limit is starting.
\item \textbf{Diffusion Policy, trimmed (1,256 to 75 passes).} The original real-robot recipe trains for about 600 passes \citep{chi2023diffusion}, so the 100- and 200-demonstration models had a quarter and an eighth of that schedule. Under-training is plausible. But success at 100 demonstrations is lower than at 200, which simple under-training does not predict, and Section~\ref{sec:4.5} shows the dip is within seed variability.
\item \textbf{$\pi_{0.5}$, LoRA.} Probably under-fitted by the recipe: the one learning-rate pilot scored 5 of 10 where the standard recipe scored 0 (Section~\ref{sec:4.4}).
\end{itemize}
\subsection{What can and cannot be concluded}\label{sec:5.3}
Three statements survive the confound, because they concern the low-data end, where exposure was highest, or compare recipes at one budget:

\begin{enumerate}
\item Below 50 demonstrations, fine-tuned $\pi_{0.5}$ far outperforms the policies trained from scratch.
\item ACT needs between 50 and 100 demonstrations to learn this task under its default recipe.
\item The initiation failure and the LoRA failure are properties of the data and the recipe, and appear at every budget.
\end{enumerate}
Two statements do not survive it and we do not make them:

\begin{enumerate}
\item That more data stopped helping $\pi_{0.5}$ after 25 demonstrations, or ACT after 100.
\item That ACT is better than $\pi_{0.5}$ at 100 and 200 demonstrations in any sense beyond ``under these two recipes''.
\end{enumerate}
Two follow-up studies address this directly (Section~\ref{sec:8}). Validation loss on the 25 held-out demonstrations, computed at each saved checkpoint, would show whether the 100- and 200-demonstration models were still improving when training stopped. Models trained with the number of passes over the data held constant across budgets would remove the confound altogether.

\FloatBarrier
\section{Discussion}\label{sec:6}
\paragraph{Pretraining buys data efficiency; on this task we did not see it buy a higher ceiling.} With 12 or 25 demonstrations, fine-tuned $\pi_{0.5}$ was far ahead of every other policy: at 12 demonstrations it reached the ball in 27 of 40 episodes, when ACT, trained on the same demonstrations, did so in 6 of 60. That advantage is what pretraining is expected to provide, and it is large. But it did not extend to a higher ceiling. From 100 demonstrations a policy with no pretraining was at least as successful, started as reliably and held a higher control rate. Earlier comparisons on low-cost arms, made at a single data size or with fewer trials, found the pretrained models ahead (Section~\ref{sec:2}); our result suggests that where a comparison sits on the data axis can decide its outcome.

\paragraph{Why $\pi_{0.5}$ stops improving is the main open question.} Its reaching is solved by 25 demonstrations and its failures are in grasping and carrying. Three explanations fit the data and predict different things. If the 6,000-step schedule under-trains the larger budgets, longer training will lift the plateau. If the frozen vision-language model limits fine grasp control, training more of the model will. If the limit is physical, in the gripper and the ball, every policy should meet it, and ACT at about 90\% suggests it is not the binding constraint. The follow-up studies in Section~\ref{sec:8} are designed to separate these.

\paragraph{The initiation failure comes from the data, and which policies inherit it is not yet explained.} The demonstrations open with a median of 2.8 seconds in which nothing moves. One relevant quantity is how far ahead a policy predicts. Diffusion Policy and SmolVLA predict 48 and 50 steps, about 1.6 seconds, which is shorter than the typical lead-in: for the first frames of most demonstrations, the entire training target is to stay still. ACT predicts 100 steps, 3.3 seconds, so most of its targets from the start pose include the departure. That account fits three of the policies and fails for the fourth: $\pi_{0.5}$ predicts 50 steps and never froze, under either recipe. \citet{chi2023diffusion} argued that predicting action sequences makes a policy robust to idle actions; our result suggests that this protection may not hold when the idle stretch is longer than the sequence. Whatever the full mechanism, the remedy is cheap. Trimming the lead-in took Diffusion Policy from starting in 4\% of episodes to starting in 90\%.

\paragraph{How a large policy is fine-tuned can matter more than how much data it gets.} The difference between the two $\pi_{0.5}$ recipes, 61 percentage points, is larger than the difference between 12 and 200 demonstrations for any policy except ACT. The failed recipe trained 0.07\% of the model's weights and the working one 16.7\%. A reader who had run only the LoRA recipe would have concluded that $\pi_{0.5}$ does not work on this arm. Negative results for large pretrained policies should be read with the recipe in view, and papers that compare such policies should report it in full.

\paragraph{Success rate is a blunt instrument at this sample size.} Outcomes on a given placement were only weakly repeatable, re-tests moved by 2 in 10, and cells of 40 episodes resolve differences of about 25 points. Three things made the data more informative at no extra bench time. Staging each episode from the recordings turned one bit per episode into an ordered outcome and located each policy's losses. Continuous measures, the accuracy of the first grasp and the agreement between successive action chunks, moved more steadily with data than success rates did. And checking the scoring sheet against the recordings found errors in 2\% of episodes, enough to move individual cells by a success or two.

\paragraph{Practical guidance for a new task on this class of arm.} These follow from one task and should be taken as starting points.

\begin{itemize}
\item With fewer than about 50 demonstrations, fine-tune a pretrained policy, and train the action expert, not low-rank adapters, unless the adapter recipe has been tuned.
\item With 100 or more demonstrations, train ACT from scratch as a baseline before assuming a larger model is needed.
\item Remove idle frames from the start of demonstrations, or start recording already in motion.
\item In evaluation, log whether the arm started and how long it took. A policy that does not start has a different problem from one that starts and fails.
\item Do not compare two models on ten episodes each unless the expected difference is very large.
\end{itemize}
\FloatBarrier
\section{Limitations}\label{sec:7}
Section~\ref{sec:1.1} stated the limits that shape the headline claims. The full list follows.

\begin{itemize}
\item \textbf{One task, one arm, one operator.} A single pick-and-place task with ten test placements that lie within the area covered by the demonstrations. Nothing here tests generalisation to new objects, positions or scenes.
\item \textbf{Fixed training length.} Data budget and training exposure are confounded within every recipe (Section~\ref{sec:5}). This is the most important limit on the high-data results.
\item \textbf{One recipe per policy.} No hyperparameters were tuned. The $\pi_{0.5}$ schedule is short relative to published configurations, and the Diffusion Policy horizon differs from both the original paper and the library default.
\item \textbf{Few and unequal seeds.} ACT and SmolVLA have four seeds per budget. Diffusion Policy on trimmed demonstrations has two. $\pi_{0.5}$ expert-only has three models from two seeds, pooled across EMA settings. The LoRA grid was stopped early at 100 and 200 demonstrations, and half of the untrimmed Diffusion Policy models were never tested.
\item \textbf{Policies were tested on different days in different light.} ACT ran at 223 to 394 lux and $\pi_{0.5}$ expert-only at 117 to 209. Brighter light helped Diffusion Policy start and did not help SmolVLA. Its effect on ACT was measured on one checkpoint (9 of 10 dim, 7 of 10 bright) and on $\pi_{0.5}$ not at all, so the ACT and $\pi_{0.5}$ comparison carries an unmeasured lighting difference that, if anything, disadvantages $\pi_{0.5}$.
\item \textbf{Testing was dimmer than training.} All main results are under an illuminance shift of roughly half.
\item \textbf{Unblinded scoring.} The operator knew which model was running. The audit against recordings corrects the errors it can detect: whether the arm moved, whether the ball was held, whether it left the table.
\item \textbf{The 30-second limit.} It turns slow successes into failures. Nine episodes ended with the ball in the gripper, and SmolVLA, the slowest to start, also lost three episodes to late starts.
\item \textbf{Automatic measures.} The split of grasp failures into location and grip failures rests on colour thresholding in the wrist view and has not yet been checked by eye on a sample.
\item \textbf{Placement consistency.} The camera data suggest that in two blocks the ball layouts ran one placement ahead for several episodes. Block totals are unaffected; comparisons by placement are.
\item \textbf{Final checkpoints only.} No checkpoint selection was done, so a recipe whose best checkpoint comes earlier in training is under-represented.
\item \textbf{Small supplementary probes.} The learning-rate pilot, the real-time chunking probe and the EMA comparisons rest on 10 to 80 episodes each.
\end{itemize}
\FloatBarrier
\section{Future work}\label{sec:8}
Five follow-up studies are planned, each on a gap this paper leaves open: whether a loss computed offline predicts real success, what the scaling curves look like when training exposure is held constant, whether a simulated rig reproduces the real results, whether simulated data can stand in for teleoperation, and how much of the measurement noise seen here an automated bench would remove.

\subsection{Training loss, validation loss and real success}\label{sec:8.1}
The first study tests whether a loss computed offline predicts success on the robot. Every model in this study has a training-loss history, saved intermediate checkpoints and a measured real-world success rate, and 25 demonstrations were held out for this purpose and never used for training (Section~\ref{sec:3.3}). Validation loss will be computed on them for every checkpoint and compared with success across budgets, seeds and policies, with particular attention to the cells suspected of under-training in Section~\ref{sec:5}. A proper test needs intermediate checkpoints run on the robot as well as final ones. If offline loss proves a usable proxy, checkpoint selection and stopping rules can be set without bench time. If it does not, as informal reports for ACT and Diffusion Policy in simulation suggest \citep{barannikov2024validation}, that is worth establishing on real hardware.

\subsection{Training with exposure held constant}\label{sec:8.2}
The second study retrains the grid with the number of passes over the data held constant across budgets, so that training steps grow in proportion to the data and the amount of data is the only thing that changes between cells. This removes the confound of Section~\ref{sec:5} and tests directly whether the plateaus of $\pi_{0.5}$ after 25 demonstrations and of ACT after 100 are ceilings or under-training. It also makes training cost a function of the data budget, which is the basis a cost comparison between policies needs.

\subsection{Simulated against physical success rates}\label{sec:8.3}
The third study measures how well a simulated copy of the rig reproduces the real results. The 86 checkpoints of the main comparison, each with a measured success rate and a staged failure profile, form a calibration set. A simulated twin of the workspace will run the same checkpoints on the same placements, and simulated success and failure stage will be compared with the real ones, in the manner of \citet{li2024simpler}. Where the two agree, simulation can supply hundreds of episodes per checkpoint and lift the sample-size limits of Section~\ref{sec:4.5}.

\subsection{Simulated reinforcement learning against teleoperated demonstrations}\label{sec:8.4}
The fourth study tests whether training data generated in simulation can replace or supplement teleoperation. The same four policies will be trained on episodes produced in simulation by a reinforcement-learning agent and compared, on the physical arm, with the models trained on teleoperated demonstrations in this paper. With the demonstration budget no longer limited by operator time, the scaling curves can be extended well beyond 200 episodes. The comparison will also show whether machine-generated demonstrations avoid the idle lead-in that human teleoperation introduced.

\subsection{Automating the test bench}\label{sec:8.5}
The fifth study automates the test bench. Every episode here needed an operator to place three balls, start the run, judge the outcome and reset, which fixed the block size at ten and made repeated tests costly. The wrist- and front-camera measures developed for the audit already recover success, start and stage from the recordings with few disagreements. Closing the loop means automatic outcome detection, automatic reset of the objects and unattended runs, in the direction of the parallel low-cost evaluation cells of \citet{selvaraj2026armnetbench}. The aim is on the order of a hundred episodes per checkpoint, with scoring that is blind by construction, which would show how much of the measurement noise in Section~\ref{sec:4.5} is a matter of bench time.

\paragraph{Smaller checks that follow directly from this paper.} Retraining SmolVLA on the trimmed demonstrations would test whether its initiation failure has the same cause as Diffusion Policy's. Two more seeds of Diffusion Policy at 50, 100 and 200 demonstrations would settle whether its dip is real.

\FloatBarrier
\section{Conclusion}\label{sec:9}
On one pick-and-place task on a low-cost arm, the answer to ``which policy, and how many demonstrations'' depended on where one stands on the data axis. With a few dozen demonstrations, a pretrained vision-language-action model with its action expert fine-tuned was far ahead of the others. With a hundred or more, a small transformer trained from scratch did at least as well. Two of the four policies often failed to start at all, a failure that removing an idle lead-in from the demonstrations cured in one and that plausibly has the same origin in the other. And for the largest model, the choice of which weights to train moved success from 4\% to 65\%.

These results hold for fixed, near-default training recipes, and the paper has tried to be exact about what that leaves open: whether the plateaus at the high-data end are ceilings or the mark of fixed training length. The recordings made it possible to say where each policy fails and how reliable each number is, at no extra bench time. We suggest that staged outcomes, re-tests of the same checkpoint and an audit of the scoring become routine in real-robot studies of this size.

\FloatBarrier
\bibliography{references}

\begin{thebibliography}{35}
\providecommand{\natexlab}[1]{#1}
\providecommand{\url}[1]{\texttt{#1}}
\expandafter\ifx\csname urlstyle\endcsname\relax
  \providecommand{\doi}[1]{doi: #1}\else
  \providecommand{\doi}{doi: \begingroup \urlstyle{rm}\Url}\fi

\bibitem[Agarwal et~al.(2021)Agarwal, Schwarzer, Castro, Courville, and
  Bellemare]{agarwal2021precipice}
Rishabh Agarwal, Max Schwarzer, Pablo~Samuel Castro, Aaron Courville, and
  Marc~G. Bellemare.
\newblock Deep reinforcement learning at the edge of the statistical precipice.
\newblock In \emph{Advances in Neural Information Processing Systems 34}, pages
  29304--29320, 2021.
\newblock arXiv:2108.13264.

\bibitem[Barannikov(2024)]{barannikov2024validation}
Marina Barannikov.
\newblock Is using a validation set useful for end-to-end learning in robotics?
\newblock Hugging Face blog,
  \url{https://huggingface.co/blog/m1b/validation-loss-robotics}, December
  2024.

\bibitem[Beyer et~al.(2024)Beyer, Steiner, Pinto, et~al.]{beyer2024paligemma}
Lucas Beyer, Andreas Steiner, Andr{\'e}~Susano Pinto, et~al.
\newblock {PaliGemma}: A versatile {3B} {VLM} for transfer, 2024.
\newblock arXiv:2407.07726.

\bibitem[Black et~al.(2025{\natexlab{a}})Black, Brown, Darpinian,
  et~al.]{black2025pi05}
Kevin Black, Noah Brown, James Darpinian, et~al.
\newblock {$\pi_{0.5}$}: a vision-language-action model with open-world
  generalization.
\newblock In \emph{Proceedings of the 9th Conference on Robot Learning}, volume
  305 of \emph{PMLR}, pages 17--40, 2025{\natexlab{a}}.
\newblock arXiv:2504.16054.

\bibitem[Black et~al.(2025{\natexlab{b}})Black, Brown, Driess,
  et~al.]{black2025pi0}
Kevin Black, Noah Brown, Danny Driess, et~al.
\newblock {$\pi_0$}: A vision-language-action flow model for general robot
  control.
\newblock In \emph{Robotics: Science and Systems XXI}, 2025{\natexlab{b}}.
\newblock arXiv:2410.24164.

\bibitem[Black et~al.(2025{\natexlab{c}})Black, Galliker, and
  Levine]{black2025rtc}
Kevin Black, Manuel~Y. Galliker, and Sergey Levine.
\newblock Real-time execution of action chunking flow policies.
\newblock In \emph{Advances in Neural Information Processing Systems 38},
  2025{\natexlab{c}}.
\newblock arXiv:2506.07339.

\bibitem[Brohan et~al.(2023)Brohan, Brown, Carbajal, et~al.]{brohan2023rt1}
Anthony Brohan, Noah Brown, Justice Carbajal, et~al.
\newblock {RT-1}: Robotics transformer for real-world control at scale.
\newblock In \emph{Robotics: Science and Systems XIX}, 2023.
\newblock arXiv:2212.06817.

\bibitem[Cadene et~al.(2024)Cadene, Alibert, Soare, et~al.]{cadene2024lerobot}
Remi Cadene, Simon Alibert, Alexander Soare, et~al.
\newblock {LeRobot}: State-of-the-art machine learning for real-world robotics
  in {PyTorch}.
\newblock Software, \url{https://github.com/huggingface/lerobot}, 2024.

\bibitem[Cadene et~al.(2026)Cadene, Alibert, Capuano,
  et~al.]{cadene2026lerobot}
Remi Cadene, Simon Alibert, Francesco Capuano, et~al.
\newblock {LeRobot}: An open-source library for end-to-end robot learning.
\newblock In \emph{The Fourteenth International Conference on Learning
  Representations}, 2026.
\newblock arXiv:2602.22818.

\bibitem[Chen et~al.(2025)Chen, Brakel, Bronars, et~al.]{chen2025idling}
Annie~S. Chen, Philemon Brakel, Antonia Bronars, et~al.
\newblock Exploiting policy idling for dexterous manipulation, 2025.
\newblock arXiv:2508.15669.

\bibitem[Chi et~al.(2023)Chi, Feng, Du, et~al.]{chi2023diffusion}
Cheng Chi, Siyuan Feng, Yilun Du, et~al.
\newblock Diffusion policy: Visuomotor policy learning via action diffusion.
\newblock In \emph{Robotics: Science and Systems XIX}, 2023.
\newblock Extended version in The International Journal of Robotics Research
  44(10--11), 2025. arXiv:2303.04137.

\bibitem[Driess et~al.(2025)Driess, Springenberg, Ichter,
  et~al.]{driess2025insulating}
Danny Driess, Jost~Tobias Springenberg, Brian Ichter, et~al.
\newblock Knowledge insulating vision-language-action models: Train fast, run
  fast, generalize better.
\newblock In \emph{Advances in Neural Information Processing Systems 38}, 2025.
\newblock arXiv:2505.23705.

\bibitem[Henderson et~al.(2018)Henderson, Islam, Bachman,
  et~al.]{henderson2018matters}
Peter Henderson, Riashat Islam, Philip Bachman, et~al.
\newblock Deep reinforcement learning that matters.
\newblock In \emph{Proceedings of the AAAI Conference on Artificial
  Intelligence}, volume~32, 2018.
\newblock arXiv:1709.06560.

\bibitem[Hu et~al.(2022)Hu, Shen, Wallis, et~al.]{hu2022lora}
Edward~J. Hu, Yelong Shen, Phillip Wallis, et~al.
\newblock {LoRA}: Low-rank adaptation of large language models.
\newblock In \emph{International Conference on Learning Representations}, 2022.
\newblock arXiv:2106.09685.

\bibitem[Kim et~al.(2024)Kim, Pertsch, Karamcheti, et~al.]{kim2024openvla}
Moo~Jin Kim, Karl Pertsch, Siddharth Karamcheti, et~al.
\newblock {OpenVLA}: An open-source vision-language-action model.
\newblock In \emph{Proceedings of the 8th Conference on Robot Learning}, volume
  270 of \emph{PMLR}, pages 2679--2713, 2024.
\newblock arXiv:2406.09246.

\bibitem[Knight et~al.(2024)Knight, Kooijmans, Cadene, et~al.]{knight2024so101}
Rob Knight, Pepijn Kooijmans, Remi Cadene, et~al.
\newblock Standard open {SO-100} \& {SO-101} arms.
\newblock Hardware design, \url{https://github.com/TheRobotStudio/SO-ARM100},
  2024.

\bibitem[Kress-Gazit et~al.(2024)Kress-Gazit, Hashimoto, Kuppuswamy,
  et~al.]{kressgazit2024empirical}
Hadas Kress-Gazit, Kunimatsu Hashimoto, Naveen Kuppuswamy, et~al.
\newblock Robot learning as an empirical science: Best practices for policy
  evaluation, 2024.
\newblock arXiv:2409.09491.

\bibitem[Li et~al.(2024)Li, Hsu, Gu, et~al.]{li2024simpler}
Xuanlin Li, Kyle Hsu, Jiayuan Gu, et~al.
\newblock Evaluating real-world robot manipulation policies in simulation.
\newblock In \emph{Proceedings of the 8th Conference on Robot Learning}, volume
  270 of \emph{PMLR}, pages 3705--3728, 2024.
\newblock arXiv:2405.05941.

\bibitem[Liang and Zeger(1986)]{liang1986gee}
Kung-Yee Liang and Scott~L. Zeger.
\newblock Longitudinal data analysis using generalized linear models.
\newblock \emph{Biometrika}, 73\penalty0 (1):\penalty0 13--22, 1986.

\bibitem[Lin et~al.(2025)Lin, Hu, Sheng, et~al.]{lin2025scaling}
Fanqi Lin, Yingdong Hu, Pingyue Sheng, et~al.
\newblock Data scaling laws in imitation learning for robotic manipulation.
\newblock In \emph{International Conference on Learning Representations}, 2025.
\newblock arXiv:2410.18647.

\bibitem[Mandlekar et~al.(2021)Mandlekar, Xu, Wong,
  et~al.]{mandlekar2021robomimic}
Ajay Mandlekar, Danfei Xu, Josiah Wong, et~al.
\newblock What matters in learning from offline human demonstrations for robot
  manipulation.
\newblock In \emph{Proceedings of the 5th Conference on Robot Learning}, volume
  164 of \emph{PMLR}, pages 1678--1690, 2021.
\newblock arXiv:2108.03298.

\bibitem[Marafioti et~al.(2025)Marafioti, Zohar, Farr{\'e},
  et~al.]{marafioti2025smolvlm}
Andr{\'e}s Marafioti, Orr Zohar, Miquel Farr{\'e}, et~al.
\newblock {SmolVLM}: Redefining small and efficient multimodal models, 2025.
\newblock arXiv:2504.05299.

\bibitem[Newcombe(1998)]{newcombe1998interval}
Robert~G. Newcombe.
\newblock Interval estimation for the difference between independent
  proportions: comparison of eleven methods.
\newblock \emph{Statistics in Medicine}, 17\penalty0 (8):\penalty0 873--890,
  1998.

\bibitem[{Physical Intelligence}(2025)]{openpi2025}
{Physical Intelligence}.
\newblock openpi.
\newblock Software, \url{https://github.com/Physical-Intelligence/openpi},
  accessed 9 Oct 2026, 2025.

\bibitem[Selvaraj et~al.(2026)Selvaraj, Uttini, and
  Kuosmanen]{selvaraj2026armnetbench}
Praveen Selvaraj, Lorenzo Uttini, and Ville Kuosmanen.
\newblock {ArmnetBench} v0.1: Parallel real-world evaluation of manipulation
  policies on a low-cost arm farm, 2026.
\newblock arXiv:2607.24481.

\bibitem[Shukor et~al.(2025)Shukor, Aubakirova, Capuano,
  et~al.]{shukor2025smolvla}
Mustafa Shukor, Dana Aubakirova, Francesco Capuano, et~al.
\newblock {SmolVLA}: A vision-language-action model for affordable and
  efficient robotics, 2025.
\newblock arXiv:2506.01844.

\bibitem[Snyder et~al.(2025)Snyder, Hancock, Badithela, et~al.]{snyder2025step}
David Snyder, Asher~James Hancock, Apurva Badithela, et~al.
\newblock Is your imitation learning policy better than mine? policy comparison
  with near-optimal stopping.
\newblock In \emph{Robotics: Science and Systems XXI}, 2025.
\newblock arXiv:2503.10966.

\bibitem[Snyder et~al.(2026)Snyder, Badithela, Matni, et~al.]{snyder2026beyond}
David Snyder, Apurva Badithela, Nikolai Matni, et~al.
\newblock Beyond binary success: Sample-efficient and statistically rigorous
  robot policy comparison.
\newblock In \emph{Robotics: Science and Systems XXII}, 2026.
\newblock arXiv:2603.13616.

\bibitem[{TRI LBM Team}(2025)]{tri2025lbm}
{TRI LBM Team}.
\newblock A careful examination of large behavior models for multitask
  dexterous manipulation, 2025.
\newblock arXiv:2507.05331.

\bibitem[Vincent et~al.(2024)Vincent, Nishimura, Itkina, Shah, Schwager, and
  Kollar]{vincent2024generalizable}
Joseph~A. Vincent, Haruki Nishimura, Masha Itkina, Paarth Shah, Mac Schwager,
  and Thomas Kollar.
\newblock How generalizable is my behavior cloning policy? a statistical
  approach to trustworthy performance evaluation.
\newblock \emph{IEEE Robotics and Automation Letters}, 9\penalty0
  (10):\penalty0 8619--8626, 2024.
\newblock arXiv:2405.05439.

\bibitem[Wang et~al.(2026)Wang, Yang, Zhang, Kim, Hardy, and
  Rajpurkar]{wang2026surgery}
Xucheng Wang, Zhizhou Yang, Xiaoman Zhang, Sung~Eun Kim, Romain Hardy, and
  Pranav Rajpurkar.
\newblock Imitation learning for robot assistance in open surgery: A
  multi-policy evaluation on suture following, 2026.
\newblock arXiv:2605.28736.

\bibitem[Wilson(1927)]{wilson1927probable}
Edwin~B. Wilson.
\newblock Probable inference, the law of succession, and statistical inference.
\newblock \emph{Journal of the American Statistical Association}, 22\penalty0
  (158):\penalty0 209--212, 1927.

\bibitem[Yu and Qiu(2026)]{yu2026benchmarking}
Yi~Yu and Xinchuan Qiu.
\newblock Benchmarking vision-language-action models on {SO-101}: Failure and
  recovery analysis, 2026.
\newblock arXiv:2606.08881.

\bibitem[Zhang et~al.(2025)Zhang, Qi, and Zheng]{zhang2025experiences}
Yihao Zhang, Yuankai Qi, and Xi~Zheng.
\newblock Experiences from benchmarking vision-language-action models for
  robotic manipulation, 2025.
\newblock arXiv:2511.11298.

\bibitem[Zhao et~al.(2023)Zhao, Kumar, Levine, and Finn]{zhao2023act}
Tony~Z. Zhao, Vikash Kumar, Sergey Levine, and Chelsea Finn.
\newblock Learning fine-grained bimanual manipulation with low-cost hardware.
\newblock In \emph{Robotics: Science and Systems XIX}, 2023.
\newblock arXiv:2304.13705.

\end{thebibliography}
\FloatBarrier
\appendix
\FloatBarrier
\section{Results by seed}\label{app:A}
Successes out of ten for every block in the main comparison. Where a checkpoint was tested more than once, the runs are listed in order with arrows.

\begin{table}[t]
\centering
\footnotesize
\caption{\textbf{Successes out of ten for every block in the main comparison, by seed.} Where a checkpoint was tested more than once, the runs are listed in order with arrows.}
\label{tab:A}
\begin{tabular}{P{3.81cm}P{2.32cm}P{1.50cm}P{1.50cm}P{1.63cm}P{1.63cm}P{1.63cm}}
\toprule
Recipe & Model & 12 demos & 25 demos & 50 demos & 100 demos & 200 demos \\
\midrule
ACT & Seed 0 & 0 & 1 & 2 & 7 & 9 \\
ACT & Seed 1 & 0 & 3 & 6 & 10 & 9 \\
ACT & Seed 2 & 1 & 3 & 6 & 10 & 8 \\
ACT & Seed 3 & 0 & 4 & 4 & 10 & 9 \\
$\pi_{0.5}$, expert-only & Seed 0, no EMA & 4 $\rightarrow$ 5 & 4 $\rightarrow$ 8 & 9 $\rightarrow$ 6 & 6 $\rightarrow$ 8 & 8 $\rightarrow$ 7 \\
$\pi_{0.5}$, expert-only & Seed 0, EMA & 3 & 9 & 5 & 8 & 8 \\
$\pi_{0.5}$, expert-only & Seed 3, EMA & 4 & 7 & 7 & 8 & 6 \\
SmolVLA & Seed 0 & 0 & 1 & 3 & 3 & 1 \\
SmolVLA & Seed 1 & 1 & 1 & 9 & 4 & 3 \\
SmolVLA & Seed 2 & 0 & 2 & 3 & 5 & 4 \\
SmolVLA & Seed 3 & 1 & 0 & 6 & 2 & 7 \\
Diffusion Policy, trimmed & Seed 1 & 0 & 3 & 4 $\rightarrow$ 7 $\rightarrow$ 8 & 6 & 3 \\
Diffusion Policy, trimmed & Seed 3 & 0 & 3 & 3 $\rightarrow$ 3 & 0 $\rightarrow$ 0 & 7 $\rightarrow$ 2 \\
\bottomrule
\end{tabular}
\end{table}

ACT rows use the first ten placements. Over all placements run, the ACT counts are: seed 0, 1/20, 1/10, 6/20, 7/10, 17/20; seed 1, 0/10, 6/20, 6/10, 18/20, 19/20; seed 2, 1/20, 7/20, 6/10, 10/10, 8/10; seed 3, 0/10, 4/10, 8/20, 16/20, 9/10.

The two recipes that barely worked, by budget from 12 to 200 demonstrations. Diffusion Policy, untrimmed: seeds 0 and 3 scored 0 and 0; seeds 0 and 2 scored 0 and 1; seeds 0 and 1 scored 0 and 0; seed 1 scored 0 and seed 2 scored 0 twice; seeds 2 and 3 scored 0 and 2. $\pi_{0.5}$, LoRA: seeds 0, 2 and 3 scored 0, 1 and 0; seeds 0, 2 and 3 scored 0, 2 and 0; seeds 1, 2 and 3 scored 0, 1 and 0; seed 0 scored 0; seed 2 scored 0.

\FloatBarrier
\section{Changes made by the scoring audit}\label{app:B}
Twenty-eight episodes were reviewed after the recordings contradicted the scoring sheet or left it in doubt. Twenty-seven were changed and one was confirmed as scored.

\begin{table}[t]
\centering
\small
\caption{\textbf{Changes made by the scoring audit.}}
\label{tab:B}
\begin{tabular}{P{5.68cm}P{2.61cm}P{6.87cm}}
\toprule
Change & Episodes & Basis \\
\midrule
Success to failure & 6 & Ball still in the gripper at 30 seconds (video, and the timeout rule) \\
Success to failure & 4 & The arm never moved \\
Success to failure & 4 & The ball never left its start position, or was knocked aside and never carried \\
Success to failure & 2 & Grip failure seen on video \\
Failure to success & 6 & Ball carried and delivered, on video or in the camera data \\
Failure class changed & 5 & Arm movement and gripper data contradicted the label \\
Confirmed as scored & 1 & Video \\
\bottomrule
\end{tabular}
\end{table}

Ten of the 27 changes are five pairs of adjacent rows in which a result was entered one row off or two labels were exchanged. The per-episode list, with the original entry and the reason for each change, is in the data file.

\FloatBarrier
\section{Measures taken from the recordings}\label{app:C}
All measures are computed from the recorded joint positions, the recorded commanded actions and the two video streams. Video is analysed at 320 $\times$ 240 pixels on every third frame, ten frames per second.

\begin{itemize}
\item \textbf{Started.} The largest excursion of any arm joint from its position at the first frame is at least 60$^\circ$. In 282 episodes it was below 47$^\circ$ and in the other 1,058 it was above 108$^\circ$. The operator's label agrees in 280 of the 282.
\item \textbf{Time to start.} The first time any arm joint is 20$^\circ$ from its position at the first frame.
\item \textbf{Ball detection.} Pixels are classed as green by fixed thresholds in hue, saturation and value, and the largest connected green region is taken as the ball. The red balls are found the same way with a second set of thresholds.
\item \textbf{Ball between the jaws.} In the wrist view, the green region is within 45 pixels of a fixed point between the jaws and its area is within the range seen when the ball is held.
\item \textbf{Reached the ball.} The ball is between the jaws in at least two analysed frames around a gripper close.
\item \textbf{Carried the ball.} The ball is between the jaws for a total of two seconds or more.
\item \textbf{First grasp attempt.} The first gripper close of the episode. It is on the ball if the ball is between the jaws at that close; the miss distance is the distance in pixels from the ball to the fixed point between the jaws. The on-ball judgement uses several frames and the size of the ball in the image as well as its position, so it is not a simple threshold on the miss distance.
\item \textbf{Control rate and pause per chunk.} An episode that runs the full 30 seconds records one frame per control step, so its frame count divided by 30 is the achieved rate. The shortfall from 30 Hz, divided by the number of chunks executed, is the pause per chunk.
\item \textbf{Chunk jump.} The root-mean-square change in commanded arm joint angles between the last action of one chunk and the first action of the next. Table~\ref{tab:10} reports the median over the boundaries that fall before the first gripper close.
\end{itemize}
\FloatBarrier
\section{Supplementary blocks}\label{app:D}
The 270 episodes outside the main comparison. All ran in dim light except the bright-light battery.

\begin{table}[t]
\centering
\footnotesize
\caption{\textbf{Supplementary blocks outside the main comparison.}}
\label{tab:D}
\begin{tabular}{P{5.16cm}P{2.49cm}P{1.96cm}P{2.14cm}P{2.85cm}}
\toprule
Probe & Checkpoints & Episodes & Successes & Never started \\
\midrule
Bright light, ACT (200 demos, seed 0) & 1 & 10 & 7 & 0 \\
Bright light, Diffusion Policy untrimmed (one per budget) & 5 & 50 & 8 & 34 \\
Bright light, SmolVLA (one per budget) & 5 & 50 & 10 & 14 \\
Diffusion Policy trimmed, EMA run, averaged weights (50 and 100 demos, seeds 1 and 3) & 4 & 40 & 15 & 1 \\
Diffusion Policy trimmed, EMA run, final weights (50 and 100 demos, seed 3) & 2 & 20 & 8 & 1 \\
Diffusion Policy untrimmed, EMA run (50 demos seed 1, 200 demos seed 3) & 2 & 20 & 5 & 8 \\
Diffusion Policy untrimmed, control (50 demos, seed 1) & 1 & 10 & 1 & 9 \\
$\pi_{0.5}$ expert-only, pilot (100 demos, seed 0) & 1 & 10 & 9 & 0 \\
$\pi_{0.5}$ LoRA, ten times the learning rate (100 demos, seed 0) & 1 & 10 & 5 & 0 \\
$\pi_{0.5}$ LoRA, real-time chunking (12, 25 and 50 demos) & 3 & 30 & 1 & 0 \\
$\pi_{0.5}$ LoRA, early probes of five episodes & 4 & 20 & 2 & 0 \\
\bottomrule
\end{tabular}
\end{table}

The bright-light blocks ran at 508 to 725 lux at the front camera, on 30 Sep 2026. The EMA blocks ran on 8 and 9 Oct at 172 to 190 lux, dimmer than the blocks they are compared with (215 to 366 lux for the trimmed cells, 243 to 278 for the untrimmed).

\FloatBarrier
\section{Results by placement}\label{app:E}
Successes over episodes for each of the ten placements, all budgets pooled. For SmolVLA and Diffusion Policy only episodes in which the arm started are counted.

\begin{table}[t]
\centering
\scriptsize
\caption{\textbf{Successes over episodes for each of the ten placements, all budgets pooled.} For SmolVLA and Diffusion Policy only episodes in which the arm started are counted.}
\label{tab:E}
\begin{tabular}{P{3.13cm}P{0.86cm}P{0.86cm}P{0.86cm}P{0.86cm}P{0.86cm}P{0.86cm}P{0.86cm}P{0.86cm}P{0.86cm}P{0.86cm}P{0.86cm}}
\toprule
Recipe & N00 & N01 & N02 & N03 & N04 & N05 & N06 & N07 & N08 & N09 & p \\
\midrule
ACT & 8/20 & 9/20 & 15/20 & 11/20 & 8/20 & 9/20 & 12/20 & 11/20 & 9/20 & 10/20 & 0.53 \\
$\pi_{0.5}$, expert-only & 12/20 & 16/20 & 16/20 & 6/20 & 10/20 & 10/20 & 12/20 & 17/20 & 15/20 & 16/20 & 0.002 \\
SmolVLA & 3/8 & 8/10 & 6/11 & 2/11 & 8/11 & 4/6 & 11/14 & 6/11 & 5/9 & 3/15 & 0.008 \\
Diffusion Policy, trimmed demos & 1/14 & 5/13 & 5/15 & 4/11 & 5/13 & 5/14 & 8/15 & 6/13 & 5/12 & 5/15 & 0.56 \\
\bottomrule
\end{tabular}
\end{table}

Pooled over budgets, placement matters for the two pretrained policies and not detectably for the two trained from scratch. Pooling mixes budgets with very different success rates, which hides placement effects for ACT: a test that permutes outcomes within blocks gives p between 0.03 and 0.07 for ACT, depending on the statistic. Placement probably matters to some degree for every policy. N03 is the weak placement for $\pi_{0.5}$. 

\FloatBarrier
\section{Training configurations}\label{app:F}
Settings that differ from the library defaults at commit \texttt{8352ca62} are stated; everything else is the default. All training ran on Hugging Face Jobs in the \texttt{huggingface/lerobot-gpu} container pinned by digest.

\begin{table}[t]
\centering
\small
\caption{\textbf{Training configurations.}}
\label{tab:F}
\begin{tabular}{P{3.77cm}P{5.46cm}P{2.82cm}P{2.82cm}}
\toprule
Recipe & Settings & Hardware & Time per run \\
\midrule
ACT & Default configuration: chunk size 100, 100 actions executed per chunk, variational objective, ResNet-18 encoder with ImageNet weights, learning rate 1e-5, batch 8, 100,000 steps & A10G (small) & 4.6 h \\
Diffusion Policy & Two observation steps, horizon 48, 24 actions executed per chunk, ResNet-18 encoder with ImageNet weights, learning rate 1e-4 with cosine schedule and 500 warm-up steps, batch 64, 100,000 steps & A10G (large) & 13.4 h \\
SmolVLA & \texttt{lerobot/smolvla\_base}, action expert trained, chunk size 50, learning rate 1e-4 with cosine decay, batch 64, 20,000 steps & A100 (large) & 2.7 h \\
$\pi_{0.5}$, LoRA & \texttt{lerobot/pi05\_base}, LoRA rank 32 and alpha 32 on the library's default targets (query and value attention projections of the action expert, plus the action and time projections), 2,705,408 trainable weights; AdamW, learning rate 2.5e-5, betas 0.9 and 0.95, weight decay 0.01, gradient clipping 1.0; 1,000 warm-up steps, cosine decay to 2.5e-6 at step 6,000; batch 32; 6,000 steps; 32-bit training with gradient checkpointing; checkpoints at 3,000 and 6,000 steps; cameras mapped to the base and left-wrist slots; images 224 $\times$ 224; 50-action chunks; quantile normalisation & A100 (large) & 10.6 h \\
$\pi_{0.5}$, expert-only & As LoRA but with no adapters: the library's train\_expert\_only and freeze\_vision\_encoder options are set, so the action expert and projections are trained in full and the vision-language model is frozen; 693,422,112 trainable weights & A100 (large) & 10.6 h \\
\bottomrule
\end{tabular}
\end{table}

\paragraph{$\pi_{0.5}$ parameter counts.} The model has 4,143,404,816 weights. The LoRA recipe's 2,705,408 trainable weights are 2,506,752 in the query and value adapters of the 18 action-expert layers and 198,656 in the adapters on the four projections. The expert-only recipe's 693,422,112 are 311,427,072 in the 18 transformer layers, 116,505,600 in the 37 time-conditioned normalisation layers, 2,165,792 in the projections and 263,323,648 in the unused vocabulary output layer. Both totals match the training logs exactly. The wide-adapter pilots place rank-32 adapters on all four attention projections and all three feed-forward projections of both the language model and the action expert, 53,282,816 weights.

\paragraph{How the $\pi_{0.5}$ schedule length was set.} The schedule was shortened twice before any model had been tested on the robot: from the library's 30,000 steps to 12,000, and then to 6,000. The evidence was a LoRA pilot at 100 demonstrations whose training loss was flat from about step 4,000 to step 12,000; the motive was training time, at 6.3 seconds per step. The length was not re-examined for the expert-only recipe.

Times are medians of completed runs and did not vary with the data budget, because steps were fixed.

\end{document}